\RequirePackage{fix-cm}
\documentclass[smallcondensed]{svjour3}     
\smartqed  
\usepackage{cite}
\usepackage[dvips]{graphicx}
\usepackage{multirow,multicol} 
\usepackage[tbtags]{amsmath}
\usepackage{amsbsy}
\usepackage{amssymb}
\usepackage{amsfonts}
\usepackage{bbm}

\usepackage{graphicx}
\usepackage[caption=false]{subfig}
\usepackage{pstricks}
\usepackage{pst-node}
\usepackage{pstricks-add}
\usepackage{url}
\usepackage{textcomp}
\usepackage{algorithmic}
\usepackage{algorithm}
\usepackage{balance}

\usepackage{verbatim}
\usepackage{indentfirst}
\usepackage{amsmath}
\usepackage{gensymb}

\usepackage{array}

\usepackage{textcomp}

\newcommand{\etal}{\textit{et al}. }
\newcommand{\ie}{\textit{i}.\textit{e}., }
\newcommand{\eg}{\textit{e}.\textit{g}., }

\usepackage{makecell}

{\begin{list}               
    {$\bullet$ \hfill}{
        \setlength{\leftmargin}{\parindent}
        \setlength{\parsep}{0.04\baselineskip}
        \setlength{\itemsep}{0.5\parsep}
        \setlength{\labelwidth}{\leftmargin}
        \setlength{\labelsep}{0em}}
    }
{\end{list}}

\providecommand{\eref}[1]{\eqref{#1}}  
\providecommand{\cref}[1]{Chapter~\ref{#1}}
\providecommand{\sref}[1]{Section~\ref{#1}}
\providecommand{\fref}[1]{Fig.~\ref{#1}}
\providecommand{\tref}[1]{Table~\ref{#1}}

\providecommand{\abs}[1]{\lvert#1\rvert}
\providecommand{\norm}[1]{\lVert#1\rVert}

\begin{document}

\title{Dense Depth Estimation from Multiple 360-degree Images Using Virtual Depth 
\thanks{This work was partially supported by the National Research Foundation of Korea~(NRF) grant funded by the Korea government~(MSIT)~(No.2020R1F1A1075428) and Institute of Information \& Communications Technology Planning \& Evaluation~(IITP) grant funded by the Korea government~(MSIT)
(No.2020-0-00994, Development of autonomous VR and AR content generation technology reflecting usage environment). }
}


\author{Seongyeop~Yang         \and
       Kunhee~Kim        
       \and Yeejin~Lee
}

\institute{S.~Yang\at
              Department of Electrical and Information Engineering, Seoul National University of Science and Technology, Seoul, 01811, South Korea. \\
              \email{syyang@seoultech.ac.kr}           
           \and
           K.~Kim \at
              Department of Electrical and Information Engineering, Seoul National University of Science and Technology, Seoul, 01811, South Korea. \\
              \email{kun0906@seoultech.ac.kr} 
           \and
           Y.~Lee \at
              Department of Electrical and Information Engineering, Seoul National University of Science and Technology, Seoul, 01811, South Korea. \\
              \email{yeejinlee@seoultech.ac.kr} 
}

\date{Received: date / Accepted: date}

\maketitle

\begin{abstract}
In this paper, we propose a dense depth estimation pipeline for multiview 360\degree\: images. The proposed pipeline leverages a spherical camera model that compensates for radial distortion in 360\degree\: images. The key contribution of this paper is the extension of a spherical camera model to multiview by introducing a translation scaling scheme. Moreover, we propose an effective dense depth estimation method by setting virtual depth and minimizing photonic reprojection error. We validate the performance of the proposed pipeline using the images of natural scenes as well as the synthesized dataset for quantitive evaluation. The experimental results verify that the proposed pipeline improves estimation accuracy compared to the current state-of-art dense depth estimation methods.
\end{abstract}

\section{Introduction}
\label{sec:introduction}
Dense depth estimation is an essential process that inferences scene structure from two-dimensional~(2D) images. It has been a major topic of research in computer vision for the past few decades due to the significant role it plays in various machine vision applications, including scene reconstruction~\cite{da2019dense, kim20133d, ma20153d, pathak20163d, pathak2016dense, schonbein2014omnidirectional, Yang_2019_CVPR, Fernandez_RAL_2020, Fangzheng_CSVT_2021}, view synthesis~\cite{lie2017key, jiang2017light}, navigation~\cite{yang2020d3vo, zhan2020visual, xue2019beyond}, and tracking~\cite{carrio2018drone, kart2019object}.

In recent years, 360\degree\: cameras have gained more popularity in applications that require a wide field of view~(FoV) such as human-robot-interaction, intelligent vehicles, surveillance, virtual reality, and augmented reality. Unlike ordinary (perspective) cameras, typical 360\degree\: cameras are equipped with fisheye lenses covering wide angles of view~(usually, $>180$\degree\:), and field of view can substantially increase using 360\degree\: cameras. However, the captured images have non-rectilinear appearances, yielding directionally dependent distortions. Due to such distortion, dense depth estimation from 360\degree\: images remains a challenge, although remarkable progress has been made in dense depth estimation from the images captured by perspective cameras~\cite{hirschmuller2007stereo, chen2001fast, wang2020fadnet, gao2020compact}. 

To extract depth from different view images, the images should be registered into one reference coordinate system using a camera model and image geometry. Due to the abovementioned radial distortion of 360\degree\: cameras, conventional camera models and image geometry with perspective projection cannot be directly applied to them; thereby, previous studies developed the camera models to depict the relationship between 3-dimensional~(3D) points based on spherical projection~\cite{geyer2000unifying, ying2004can, courbon2007generic, li2008binocular, pagani2011structure, ma20153d, kim20133d, im2016all, da2019dense}. Also, to accurately estimate relative camera positions from different views, the work in \cite{da2019dense} investigated a way to obtain better dense correspondences using optical flow~\cite{weinzaepfel2013deepflow} along with visual features~(\eg SIFT~\cite{zhao2015sphorb}, SPHORB~\cite{rublee2011orb}, etc.), similar to the works in \cite{pathak2016dense, schonbein2014omnidirectional}. 

Toward 3D dense reconstruction from spherical images, Arican and Frossard~\cite{arican2007dense} obtained disparity maps from two views by optimizing a pixel-wise energy function using graph cuts. The work in \cite{schonbein2014omnidirectional} estimated disparity based on the plane-based prior models from two temporally and two spatially views. Kim and Hilton~\cite{kim20133d} used a partial differential equation-based disparity estimation algorithm under the assumption that image pairs were acquired in the limited configuration -- stereo spherical image pairs aligned with a vertical displacement between camera views like the setting proposed in \cite{li2008binocular}. Im \etal~\cite{im2016all} estimated a dense depth map using the spherical sweeping algorithm inspired by the plane sweeping algorithm~\cite{collins1996space}.
Owing to the development of deep learning frameworks in stereo depth estimation~\cite{chang2018pyramid, kendall2017end, Zhang_2019_CVPR}, the recent work in \cite{wang2020360sd} employed a convolutional neural network framework based on two camera configurations in a top-bottom manner~\cite{li2008binocular, kim20133d}, showing satisfactory results. Nevertheless, these works were developed in limited camera configurations, making them restricted to applying unarranged settings.
Furthermore, Zioulis \etal~\cite{zioulis2019spherical} designed a self-supervised learning framework that synthesis three views using estimated disparity maps from the convolutional neural network. More recently, the framework to fuse features from equirectangular and cubemap projections was proposed to estimate depth in monocular 360\degree\: images~\cite{Wang_2020_CVPR, 9353978}. Although the framework only requires an image to estimate depth, additional information such as cube map projection is necessary, and the estimation performance is not guaranteed.

\begin{figure}[!t]
    \centering
    \captionsetup{justification=centering}
    \includegraphics[scale=0.22]{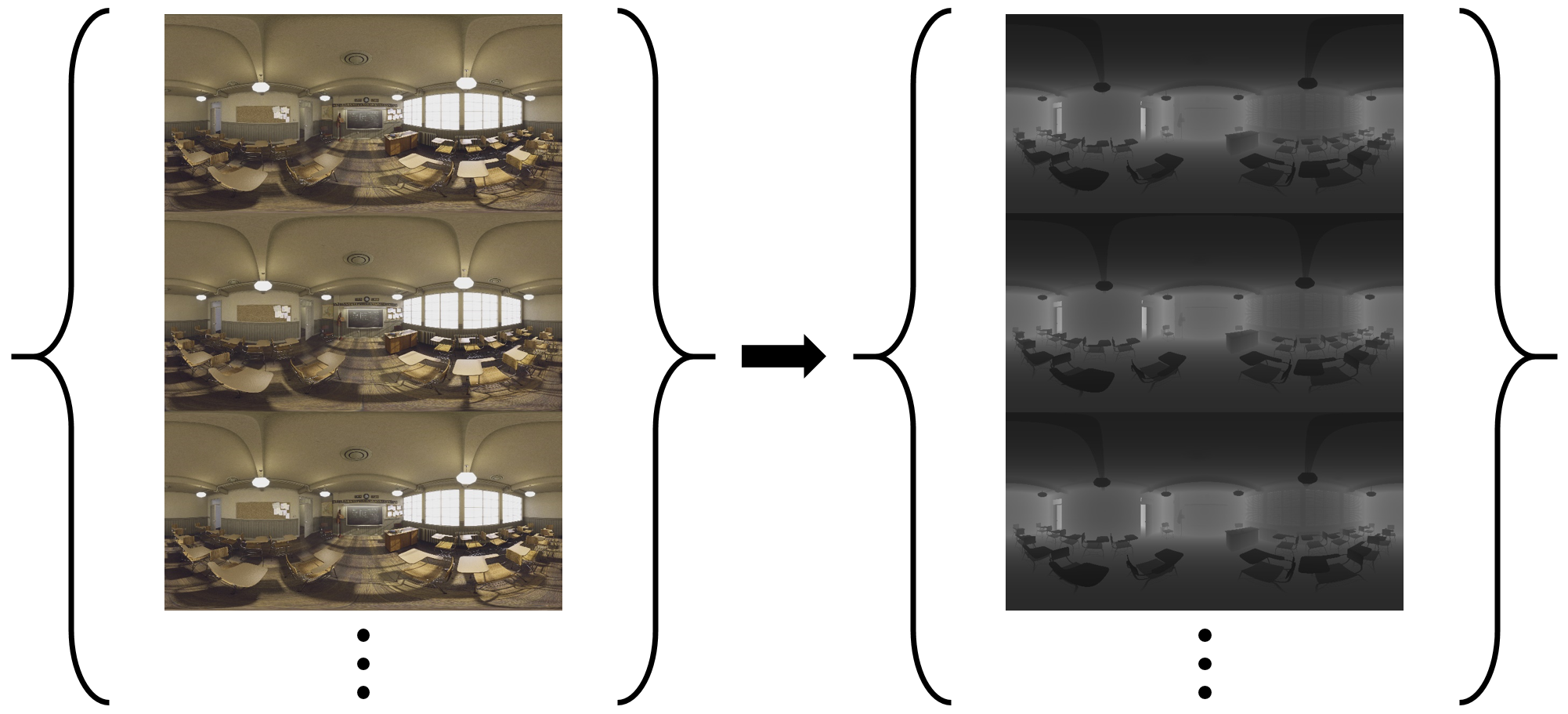}
    \caption{Given multiple 360\degree\: images, this works aims to estimate dense depth maps. The proposed pipeline transforms equirectangular images into their corresponding dense depth maps. \vspace{-0.1cm}}
    \label{fig:intro}
\end{figure}

Although the previous researches have provided decent results, most approaches have been restricted to either two-view analysis or limited camera configurations. This indicates that typical spherical camera models are inadequate and possibly inferior to multiview settings. Consequently, in this paper, we develop a practical pipeline that estimates dense depth maps from multiple 360\degree\: images, as described in \fref{fig:intro}. We expand the two-view spherical camera model to multiview and estimate camera poses by taking scaling into account. The proposed pose estimation employs a scaling factor to estimate actual relative translation between cameras, in contrast to that typical pose estimation methods find solutions with unknown scale. The camera coordinates are then registered based on the estimated poses. Finally, dense depth maps are estimated by setting virtual depth and finding the optimal depth that minimizes reprojection error in corresponding regions.   

The remainder of this paper is organized as follows. In \sref{sec:method}, we develop the proposed dense depth estimation pipeline for multiview 360\degree\: images. Then, experimental results are shown in \sref{sec:results}, and concluding remarks are made in \sref{sec:conclusion}. 

\section{Proposed Dense Depth Estimation Pipeline}
\label{sec:method}
\begin{figure*}[!t]
    \centering
    \captionsetup{justification=centering}
    \includegraphics[scale=0.46]{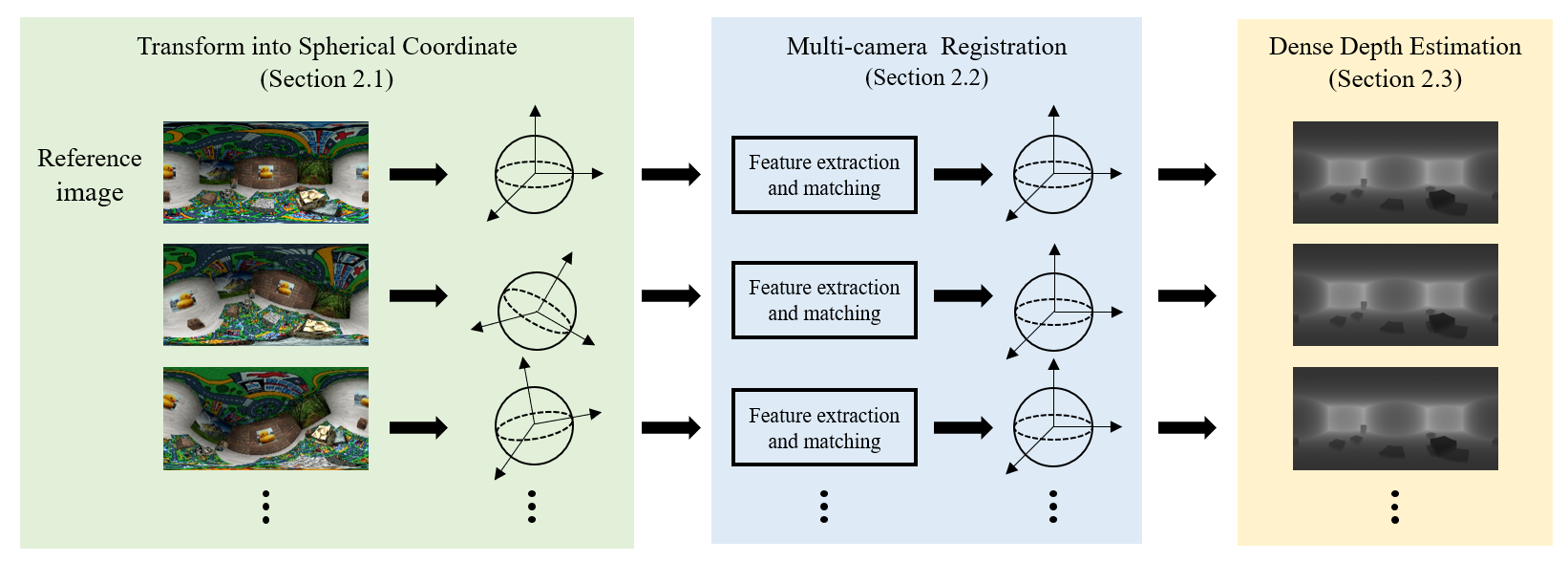}
    \caption{Overview of the proposed pipeline. The proposed pipeline consists of three steps: coordinate transformation, multi-camera registration, and dense depth estimation. \vspace{0.3cm}}
    \label{fig: overview}
\end{figure*}

The proposed dense depth estimation pipeline is composed of three steps, as shown in \fref{fig: overview}. As the first step, we transform the input image coordinate to the new coordinate by the spherical camera model in \sref{sec:spherical_camera_model}. The geometrically transformed images are then registered with the related position to the reference image as described in \sref{sec:pose_estimation}. After the registration, the dense depth map for each view is computed by back-projecting to virtual depth in \sref{sec:depth_estimation}.    

\begin{figure*}[!t]
    \centering
    \captionsetup{justification=centering}
    \includegraphics[scale=0.85]{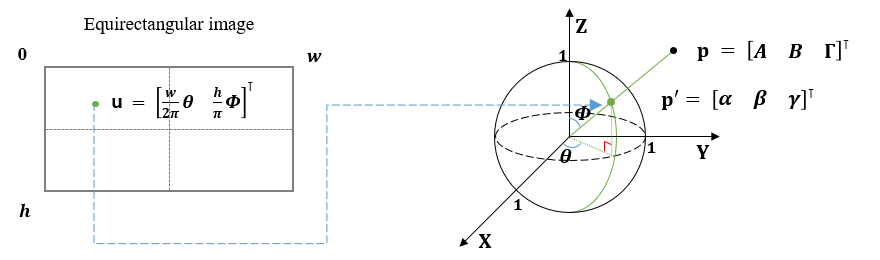} \vspace{0.1cm}
    \caption{Spherical camera model. The point $\mathbf{p}$ in three-dimensional space is projected onto the unit sphere surface, resulting in a point $\mathbf{p}'$, which corresponds to the point $\mathbf{u}$ in the equirectangular image plane. \vspace{0.2cm}}
    \label{fig: camera model}
\end{figure*}

\subsection{Spherical Camera Model}
\label{sec:spherical_camera_model}
The inputs of the proposed pipeline are equirectangular images as shown in \fref{fig:intro} and \fref{fig: overview}. To represent the input images in three-dimensional space, we employ a camera model based on spherical projection, as described in \fref{fig: camera model}. The projection model provides a convenient way to interpret the equirectangular image in three-dimensional space, making it easily compute the geometric relationship between multiple views.  

Let $\mathbf{p} = [A\; B\; \Gamma]^\intercal \in \mathbb{R}^{3}$ be a point in 3D space, and the point $\mathbf{p}'= [\alpha\; \beta\; \gamma]^\intercal \in \mathbb{R}^{3}$ be the one that $\mathbf{p}$ is projected onto a spherical surface. The point $\mathbf{p}^\prime$ can be also represented as a unit vector $\mathbf{p}/\norm{\mathbf{p}}$ associated with $\mathbf{p}$ when it is projected onto the unit sphere~(\ie $r=1$), where $\norm{\cdot}$ denotes the $L2$-norm, and $r$ is the radius of the sphere. 

By rewriting $\mathbf{p}'$ in spherical coordinate, $\mathbf{p}$ is transformed into $\mathbf{p}'=$ \\ $\left[ r\cos\theta\sin\phi \;\;\;  r\sin\theta\sin\phi \;\;\;  r\cos\phi \right]^\intercal$, where $\phi \in [0, \pi)$ is the polar angle from the positive direction of the $Z$-axis to the straight line formed by the origin and the point $\mathbf{p}'$; $\theta \in [0, 2\pi)$ is the azimuth angle from the positive direction of the $X$-axis to the straight line formed by the origin and the point $\mathbf{p}$.  

Transforming $\mathbf{p}'$ into image coordinate, the point corresponds to the longitude and the latitude on the unit sphere. The corresponding pixel location $\mathbf{u} = [u\;\; v]^\intercal \in \mathbb{R}^2$ is then represented as $\mathbf{u} = [\frac{w}{2\pi}\theta \;\; \frac{h}{\pi}\phi]^\intercal$, where $w$ and $h$ are the width and height of the equirectangular image, respectively. In the proposed pipeline, given pixel locations $\mathbf{u}$, we map them into the points $\mathbf{p}'$ on the unit sphere such that $\mathbf{u} \mapsto \mathbf{p}'$.

\subsection{Multi-camera Registration}
\label{sec:pose_estimation}
Image registration is to transform different view images into one reference coordinate by estimating relative positions between views. The relative position to the reference view can be obtained by using the epipolar geometry and by finding the mapping between corresponding points in different view images. 

Given $N$ input images, let two points $\mathbf{p}'_i = [x_i'\;\; y_i'\;\; z_i']^\intercal$ and $\mathbf{p}'_j = [x_j'\;\; y_j'\;\; z_j']^\intercal$ be the projection of the point $\mathbf{p}$ onto the $i$-th and $j$-th unit sphere centered at $O_i$ and $O_j$, respectively. Then, the purpose of image registration is to estimate the mapping from $\mathbf{p}'_i \mapsto \mathbf{p}'_j$, which is represented by camera matrices. 

A camera matrix $\mathbf{C}$ is decomposed as calibration parameters and orientation/translation, as follows: 
\begin{align}\label{eq:camera_matrix}
\mathbf{C} = \mathbf{K}\left[ \: \mathbf{R}\: | \: \mathbf{t} \:\right],  
\end{align}
where $\mathbf{K}$ is a calibration matrix, $\mathbf{R}$ is a rotation matrix, and $\mathbf{t}$ is a translation vector. Since a spherical camera model does not require intrinsic calibration, the model in \eref{eq:camera_matrix} is simplified as the normalized camera matrix having an identity calibration matrix~(\ie $\mathbf{K} = \mathbf{I}$), which removes the effect of the calibration matrix~\cite{guan2016structure, ma20153d}. Considering a pair of normalized camera matrices $\mathbf{C}_i = \left[ \:\mathbf{I}\: | \:\mathbf{0}\: \right]$ for the reference view $i$ and $\mathbf{C}_{ij} = \left[ \:\mathbf{R}_{ij}\: |\: \mathbf{t}_{ij}\:\right]$ for the view $j$ with respect to the reference view $i$, the essential matrix $\mathbf{E}_{ij}$ for the $i$-th and $j$-th views has the below form, based on the framework in ~\cite{hartley2003multiple}:
\begin{align}\label{eq:essential_matrix_1}
    \mathbf{E}_{ij} =  \begin{bmatrix} \mathbf{t}_{ij} \end{bmatrix}_\times \mathbf{R}_{ij} = \mathbf{R}_{ij} \begin{bmatrix} \mathbf{R}_{ij}^\intercal \mathbf{t}_{ij} \end{bmatrix}_\times,
\end{align}
where a matrix $\begin{bmatrix}\cdot\end{bmatrix}_\times$ is in the form of a skew-symmetric matrix. Then, the essential matrix $\mathbf{E}_{ij}$ satisfies the below condition in terms of the corresponding points $\mathbf{p}'_i \leftrightarrow \mathbf{p}'_j$:
\begin{align}\label{eq:essential_matrix_2}
    \mathbf{p}'_{j}{}^\intercal \mathbf{E}_{ij} \: \mathbf{p}'_i = 0.
\end{align}

\begin{figure*}[!t]
    \centering
    \captionsetup{justification=centering} \vspace{0.1cm}
    \includegraphics[scale=0.33]{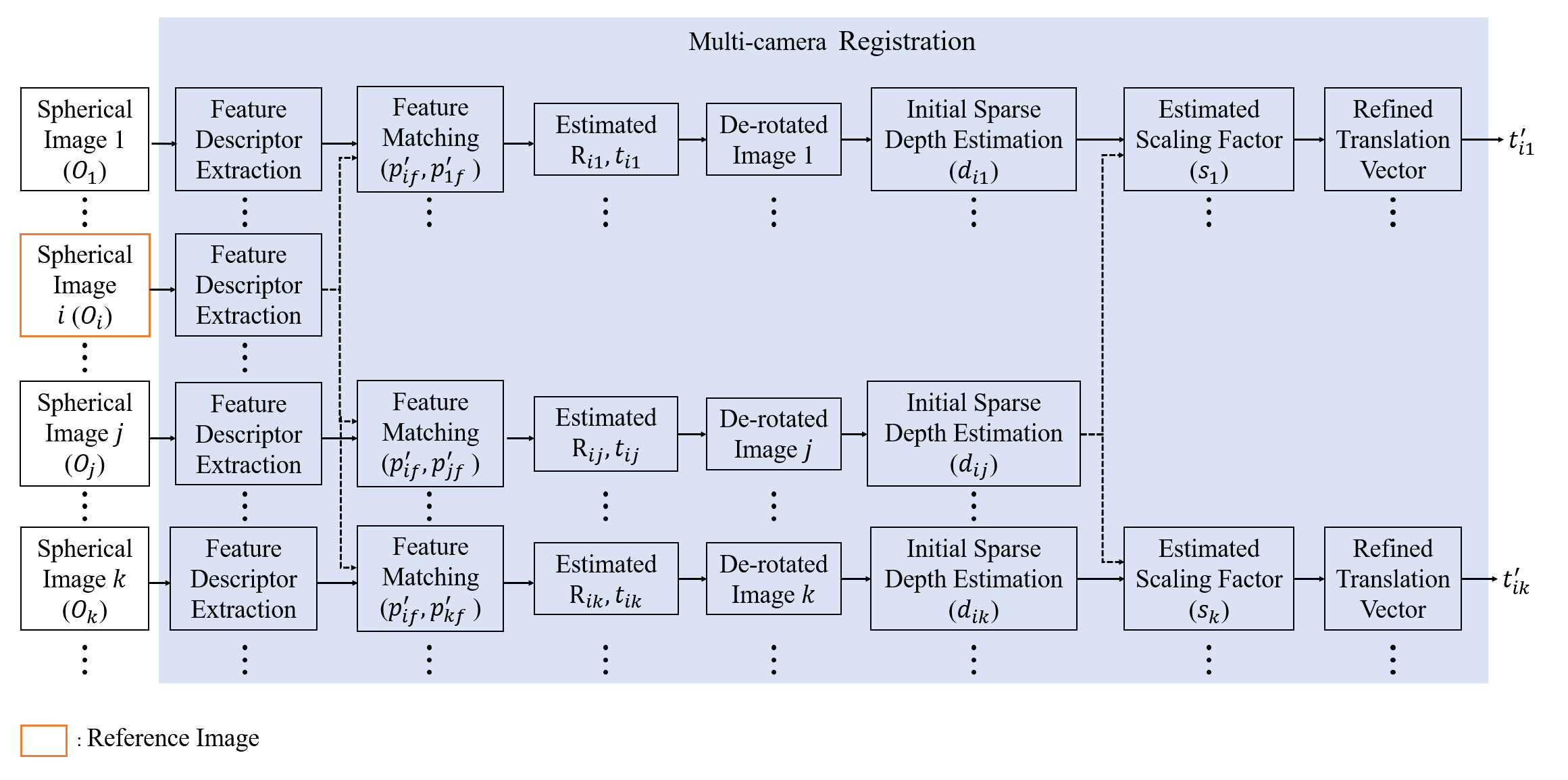}
    \caption{Description of the proposed multi-camera registration algorithm. The proposed multi-camera registration algorithm re-estimates the relative translation vectors from initial depth.} \vspace{0.2cm}
    \label{fig:block_diagram_scaling}
\end{figure*}

Given sufficient corresponding points, \eref{eq:essential_matrix_2} is used to compute the unknown matrix $\mathbf{E}_{ij}$. The matrix $\mathbf{R}_{ij}$ and the vector $\mathbf{t}_{ij}$ can then be obtained by decomposing the estimated $\mathbf{E}_{ij}$. However, the solution vector of \eref{eq:essential_matrix_2} is defined only up to an unknown scale~\cite{hartley1997defense, hartley2003multiple}. In order to avoid this trivial solution, we impose an additional constraint that the norm of the solution vector is equal to $1$. Under these conditions, it is possible to find a solution with corresponding points.

In practice, to compute $\mathbf{E}_{ij}$, we obtain feature points using the spherical oriented fast and rotated binary robust independent elementary features~(SPHORB) descriptor. It extracts points from evenly spaced grid cells on the sphere, reducing distance distortion compared to latitude-longitude images and cube maps~\cite{zhao2015sphorb, rublee2011orb, Silveira_2019_CVPR, li2018spherical}. The corresponding points are then matched by the Brute-Force matchers~\cite{2019opencv}. For those corresponding points, the 8-point algorithm is used with the Sampson approximation~\cite{pagani2011structure, hartley2003multiple} and random sample consensus~(RANSAC)~\cite{fischler1981random} to solve \eref{eq:essential_matrix_2}. Among the four possible solutions to the decomposition of \eref{eq:essential_matrix_1}, we choose the one that has the greatest amount of positive depth values because the depth values should be greater than zero.

\begin{figure}[!t]
    \centering
    \captionsetup{justification=centering}
    \includegraphics[scale=0.5]{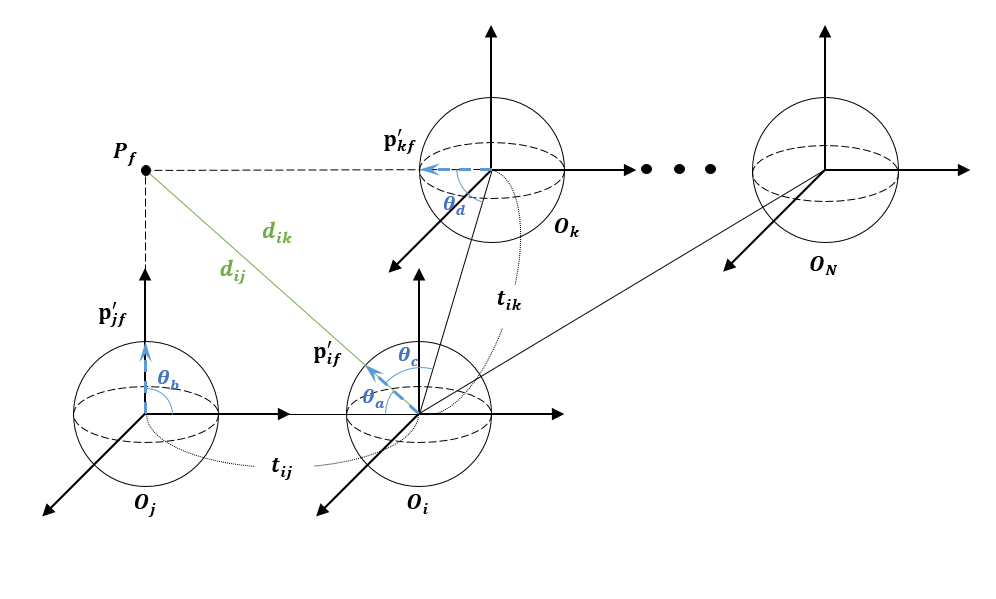}
    \caption{Multiple camera model after registration. The scaled translations $\mathbf{t}'_{ik}$ are estimated from the ratio of the initially estimated depths $d_{ij}$ and $d_{ik}$, where $d_{ij}$ is considered as a reference baseline.} \vspace{0.2cm}
    \label{fig: initial depth}
\end{figure}

Although the model of \eref{eq:essential_matrix_1} describes the relative geometric relationship between views, the actual translations between images are unable to be computed since the solution is obtained using the additional constraint. 
Therefore, in order to accurately estimate the structure of the scene from multiview, we employ a scaling factor and re-estimated the translation vector taking relative distances between views into account, as described in \fref{fig:block_diagram_scaling}. The re-estimated translation $\mathbf{t}_{ik}$ between $i$-the view and $k$-th view can be approximately measured using the scaling factor $s_k$ and the initially estimated translation $\mathbf{t}_{ik}$:
\begin{align}\label{eq:t_vector}
   \mathbf{t}_{ik}^\prime = s_k \cdot \mathbf{t}_{ik},
\end{align}
where the initial translation vector is normalized by the norm of $\norm{\mathbf{t}_{ik}} = 1$. 
We define the scaling factor $s_k$ as the ratio of the estimated depth $d_{ij}$ from the $i$-th view with respect to the $j$-th view to the estimated depth $d_{ik}$ with respect to the $k$-th view   
\begin{align}\label{eq:scaling}
    s_k = \frac{d_{ij}}{d_{ik}}.
\end{align}
The initially estimated depth $d_{ij}$ and $d_{ik}$ are computed by the inverse relationship of $\mathbf{R}^{-1}_{ij}$, $-\mathbf{t}_{ij}$, $\mathbf{R}^{-1}_{ik}$, and $-\mathbf{t}_{ik}$, respectively. The scaling factor $s_k$ can be interpreted as the measurement of relative position between the $j$-th view and the $k$-th view, where the estimated depth $d_{ij}$ and $d_{ik}$ are supposed to be equal, ideally~(See \fref{fig: initial depth}).

Defining the corresponding $f$-th feature point $\mathbf{p}_f$ of the $i$-th, $j$-th and $k$-th views as $\mathbf{p}'_{if}$, $\mathbf{p}'_{jf}$, and $\mathbf{p}'_{kf}$, the initial depth $d_{ij}$ and $d_{ik}$ are computed as 
\begin{align}\label{eq:initial_depth}
    \begin{array}{c}
    d_{ij} = \frac{\sin{\theta_b}}{\sin{\left(\theta_a + \theta_b\right)}}, 
    d_{ik} = \frac{\sin{\theta_d}}{\sin{\left(\theta_c + \theta_d\right)}} \vspace{0.3cm}, \\
    \theta_a = \cos^{-1} \left( \mathbf{p}'_{if} \cdot \mathbf{t}_{ij} \right), \:
    \theta_b = -\cos^{-1} \left( \mathbf{p}'_{jf} \cdot \mathbf{t}_{ij}\right), \vspace{0.3cm}\\
    \theta_c = \cos^{-1} \left( \mathbf{p}'_{if} \cdot \mathbf{t}_{ik} \right), \:
    \theta_d = -\cos^{-1} \left( \mathbf{p}'_{kf} \cdot \mathbf{t}_{ik} \right). \vspace{0.2cm}
    \end{array}
\end{align}
Given \eref{eq:scaling} and \eref{eq:initial_depth}, the scaling factor can be measured from different feature points. Thus, we choose a scaling factor $s_k$ as $s_{kf}$ that forms the cluster including majority neighbors within the distance $\kappa$, where $s_{kf}$ is the scaling factor measured from the $f$-th feature point. 

\subsection{Dense Depth Estimation}
\label{sec:depth_estimation}
\begin{figure*}[!t]
    \centering
    \captionsetup{justification=centering} \vspace{0.1cm}
    \includegraphics[scale=0.4]{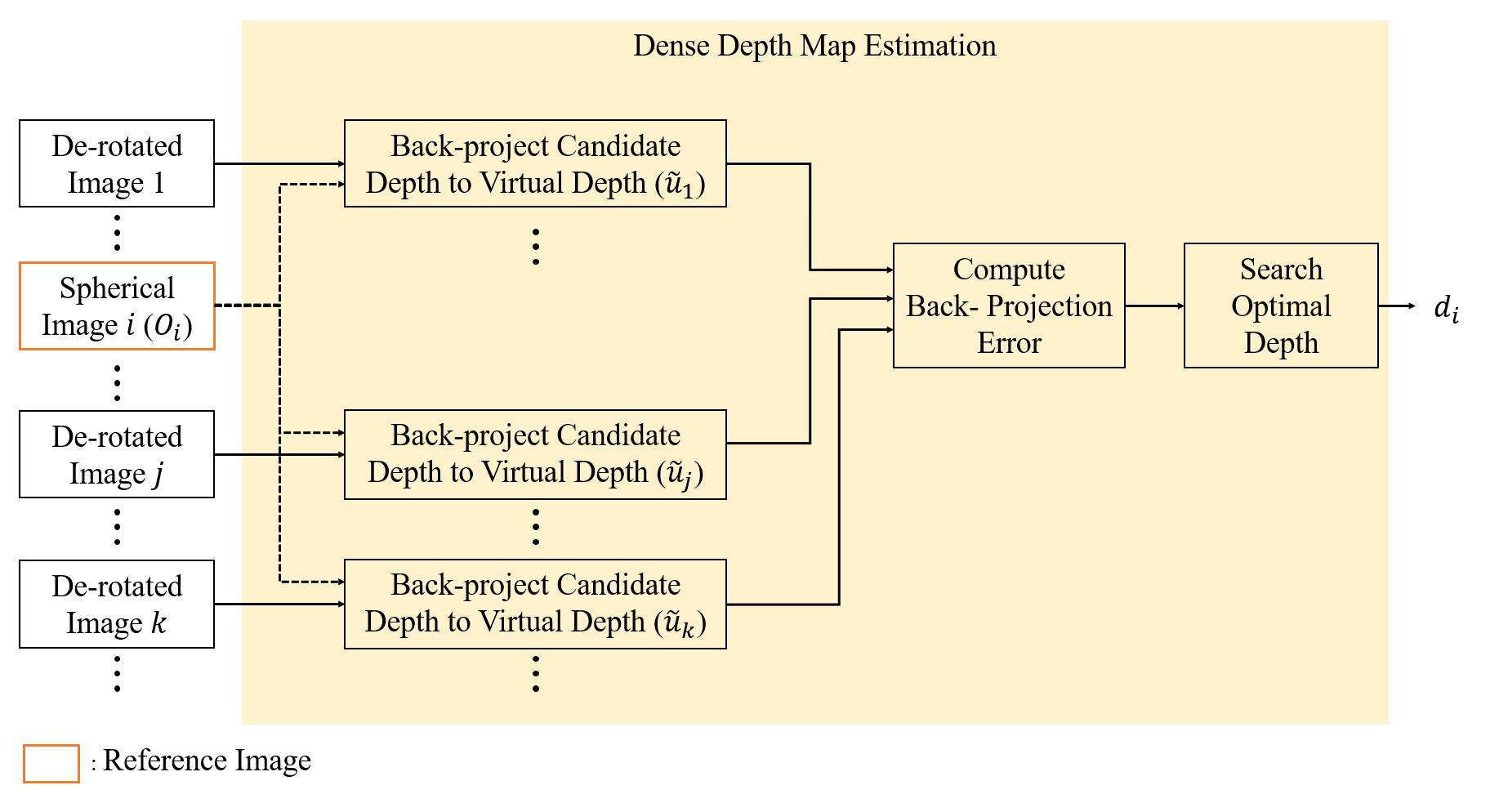}
    \caption{Description of the dense depth estimation algorithm. The proposed algorithm sets virtual depth and searches the optimal depth that minimizes reprojection error.} \vspace{0.2cm}
    \label{fig:block_diagram_depth}
\end{figure*}

After registering multiview images based on the relative positions computed in \sref{sec:pose_estimation}, we estimate dense depth maps by setting virtual depth and finding a solution that minimizes photonic reprojection error in corresponding regions directly on the unit sphere. To do this, the pixels in the local area of a reference image are first back-projected onto the target image at a certain distance. The optimal depth is then computed as the one that has the minimum pixel consistency error between the reference image and the de-rotated image from a target image among a predefined candidate set. The steps of the proposed dense depth estimation algorithm are shown in \fref{fig:block_diagram_depth}.

Define $\mathcal{F}(\cdot)$ as a function that converts image coordinate into spherical coordinate to find a point projected from virtual depth~(\ie, $\mathcal{F}\left(\mathbf{u}\right) = \mathbf{p}^\prime$). The candidate projected point onto the $j$-th viewpoint $\tilde{\mathbf{u}}_j$ of a pixel location $\mathbf{u}_i$ of the $i$-th viewpoint in a virtual depth $d'$ can be written, as follows:
\begin{align}\label{eq:convert_u}
    \begin{array}{ccl}
        \tilde{\mathbf{u}}_j & = & \emph{proj}\left( \mathbf{u}_i, d'\right) \vspace{0.2cm}\\
        & = & \mathcal{F}^{-1}\left(
        \frac{d' \cdot \mathcal{F}\left( \mathbf{u}_i\right)+\mathbf{t}^\prime_{ij}}{\norm{d' \cdot \mathcal{F}\left( \mathbf{u}_i\right)+\mathbf{t}^\prime_{ij}}} \right),
    \end{array}
\end{align}
where $\mathcal{F}^{-1}(\cdot)$ is the inverse function of $\mathcal{F}(\cdot)$~(\ie $\mathcal{F}^{-1}(\mathbf{p}^\prime)=\mathbf{u}$). The candidate set of the virtual depth can be generated by any policy. In this paper, the points are generated uniformly from the predefined minimum depth $0.05m$ and maximum depth $10.0m$.

In order to determine an optimal depth $d_i$ among the candidates, we define a cost function that minimizes reprojection error in the region $\Omega$ in the vicinity of $\mathbf{u}_i$: 
\begin{align}
    d_{i} = \underset{d^\prime}{argmin}\sum_{}^\Omega\sum_{k\notin i}^N\sum_{c} \abs{I\left(\mathbf{u}_i, c\right) - I\left(\tilde{\mathbf{u}}_k,c\right)}, 
\end{align}
where $I(\cdot)$ denotes the intensity value at the pixel location, and $c \in \{r,g,b\}$ denotes the $c$-th color channel. Intuitively, the cost function measures how well the corresponding regions are matched, which provides a way to assure pixel consistency in overlap regions. In the experiment, the reference and de-rotated images were smoothed using the bilateral filter~\cite{tomasi1998bilateral} to relieve insignificant small variations. The final depth maps were smoothed using the Fast Bilateral Solver~(FBS)~\cite{barron2016fast}.

\section{Experimental Results}
\label{sec:results}
In this section, we validated the proposed dense depth estimation method in \sref{sec:depth_estimation} as well as the pose estimation method in \sref{sec:pose_estimation} using synthesized datasets and real-world datasets. As publicly available datasets for multiview spherical images with ground-truth pose and ground-truth depth are limited, we created the synthesized dataset with 9-camera settings including benchmark datasets~(the \textit{classroom}~\cite{jeong20183dof+} and the modified \textit{smallroom}~\cite{zhang2016benefit}) for quantitative validation. In addition to quantitative evaluation, we also captured the real-world dataset using Ricoh Theta SC for qualitative validation. 

\subsection{Pose Estimation}
\label{sub: exp pose estimation}
\begin{table}[!t]
    \centering
    \caption{The estimation accuracy in terms of the average and standard deviation~($mean \pm std.$) of the estimated values of $100$ trials for each test set. GT stands for ground-truth values.}
     {\renewcommand{\arraystretch}{1.3}
    \begin{tabular}{c|@{}c@{}||@{}c@{}|@{}c@{}|@{}c@{}}
        \hline
        \multicolumn{2}{c||}{Test Set} & $\left(roll, pitch, yaw\right)$ & $t$ & $\left(s_2, s_3, s_4\right)$ \\ \hline
        \multirow{3}{*}{1} & GT & $\left(10\degree, 20\degree, 30\degree\right)$  & $\left(1, 0, 0\right)$ & $\left(0.57, 1.17, 1.73\right)$ \\ \cline{2-5}
        & \multirow{2}{*}{Ours} & $\left(9.9\degree \pm 0.09\degree, 20.0\degree \pm 0.10\degree\right., $  & $\left(1.0 \pm 0.002, 0.0 \pm 0.042, \right.$  & $\left(0.57 \pm 0.07, 1.19 \pm 0.08, \right.$ \\ 
         & & $\left. 30.1\degree \pm 0.11\degree \right)$ & $\left. 0.0 \pm 0.033\right)$ & $\left. 1.73 \pm 0.12\right)$ \\ \hline
         
         \multirow{3}{*}{2} & GT & $\left(5\degree, 12\degree, 22\degree\right)$ & $\left(0.87, -0.49, 0\right)$ & $\left(0.86, 0.86, 1.36\right)$ \\ \cline{2-5}
        & \multirow{2}{*}{Ours} & $\left(4.9\degree \pm 0.08\degree, 12.0\degree \pm 0.08\degree\right., $ & $\left(0.86 \pm 0.024, -0.50 \pm 0.042, \right.$ & $\left(0.86 \pm 0.07, 0.86 \pm 0.06, \right.$ \\ 
        & & $\left. 22.0\degree \pm 0.12\degree \right)$ & $\left. -0.1 \pm 0.022\right)$ & $\left. 1.31 \pm 0.10\right)$ \\ \hline
        
        \multirow{3}{*}{3} & GT & $\left(-3\degree, -23\degree, 30\degree\right)$ & $\left(0.5, -0.87, 0\right)$ & $\left(0.73, 0.51, 1\right)$ \\ \cline{2-5}
        & \multirow{2}{*}{Ours} & $\left(-2.9\degree \pm 0.11\degree, -23.0\degree \pm 0.09\degree\right., $ & $\left(0.51 \pm 0.034, -0.86 \pm 0.021, \right.$ & $\left(0.68 \pm 0.05, 0.48 \pm 0.05, \right.$ \\ 
        & & $\left. 30.0\degree \pm 0.12\degree \right)$ & $\left. 0.1 \pm 0.020\right)$ & $\left. 0.94 \pm 0.09\right)$\\ \hline
        
        \multirow{3}{*}{4} & GT & $\left(11\degree, -39\degree, -8\degree\right)$ & $\left(0.5, 0.87, 0\right)$ & $\left(1, 1.49, 0.86\right)$ \\ \cline{2-5}
        & \multirow{2}{*}{Ours} & $\left(11.3\degree \pm 0.17\degree, -39.0\degree \pm 0.12\degree\right., $ & $\left(0.50 \pm 0.044, 0.86 \pm 0.025, \right.$ & $\left(0.99 \pm 0.07, 1.52 \pm 0.11, \right.$ \\ 
        & & $\left. -8.2\degree \pm 0.15\degree \right)$ & $\left. 0.0 \pm 0.029\right)$ & $\left. 0.81 \pm 0.06\right)$ \\ \hline
        
        \multirow{3}{*}{5} & GT & $\left(-25\degree, 37\degree, -29\degree\right)$ & $\left(-0.61, -0.35, 0.71\right)$ & $\left(0.6, 1, 0.6\right)$ \\ \cline{2-5}
        & \multirow{2}{*}{Ours} & $\left(-25.3\degree \pm 0.12\degree, 36.9\degree \pm 0.10\degree\right., $ & $\left(-0.59 \pm 0.031, -0.35 \pm 0.047, \right.$ & $\left(0.61 \pm 0.06, 0.96 \pm 0.05, \right.$ \\ 
        & & $\left. -29.1\degree \pm 0.16\degree \right)$ & $\left. 0.73 \pm 0.018\right)$ & $\left. 0.57 \pm 0.05\right)$ \\ \hline
    \end{tabular}}
    \label{tab: pose table}
\end{table}

To verify the pose estimation performance, we created a test for unaligned images with arbitrary chosen rotation and translation. We first generated the rotated image based on model \eref{eq:camera_matrix}, as summarized in \tref{tab: pose table}. Then, the de-rotated images were reconstructed using the proposed multi-camera registration algorithm in \sref{sec:pose_estimation}, where $\kappa$ was set to $0.01$. The test was repeated $100$ times for each test set with the approximately $8500$ key points due to the randomness of the RANSAC algorithm~\cite{fischler1981random}. 

We report the pose estimation accuracy in \tref{tab: pose table} in terms of the mean and standard deviation of the estimated rotation, translation, and scaling factor of the $100$ trials for each test set. As proved in the results, the proposed pose estimation method can precisely estimate camera pose for multiview spherical images.  

\subsection{Dense Depth Estimation}
\label{sub: exp dense depth estimation}
For quantitative evaluation of the proposed dense depth estimation method in \sref{sec:depth_estimation}, we compared the performance of the proposed algorithm against three state-of-art dense depth estimation methods: the geometry and context network~(``GC-Net'')~\cite{kendall2017end}, the pyramid stereo matching network~(``PSMNet'')~\cite{chang2018pyramid}, the guided aggregation network~(``GA-Net'')~\cite{Zhang_2019_CVPR}, the 360° stereo depth estimation network~(``360SD-Net'')~\cite{wang2020360sd}, the bi-directional fusion~(``BiFuse'')~\cite{Wang_2020_CVPR}, and the unidirectional fusion~(``UniFuse'')~\cite{9353978}. Since most dense estimation algorithms were designed for two views, we chose arbitrary two cameras for the \textit{classroom} and the \textit{smallroom} sequence for the comparing algorithms. In addition to the comparing algorithms, we considered the proposed dense estimation method using known parameters and unknown pose parameters for both two- and four-camera settings. 
In the two-camera setting, we used $\mathbf{R}_{12} = \mathbf{I}$ and $\mathbf{t}_{12} = \begin{bmatrix} 0,\: & 0,\: & -1 \end{bmatrix}^\intercal$ for both the \textit{classroom} and the \textit{smallroom} sequence, where the camera $1$ was the reference. To thoroughly examine the effectiveness of the re-estimated translation vectors without any secondary effects, in the four-camera setting, we used the rotation matrix $\mathbf{R}_{ij} = \mathbf{I}$ for all cameras using the camera $1$ as reference. The translation vectors $\mathbf{t}_{12}$, $\mathbf{t}_{13}$, and $\mathbf{t}_{14}$ were set to $\begin{bmatrix} 0,\: & 0,\: & -1 \end{bmatrix}^\intercal$, $\begin{bmatrix} 0.866,\: & 0.5,\: & 0 \end{bmatrix}^\intercal$, and $\begin{bmatrix} -0.83,\: & 1.44,\: & 0 \end{bmatrix}^\intercal$, respectively, for the \textit{classroom} sequence, and $\begin{bmatrix} 0,\: & 0,\: & -1 \end{bmatrix}^\intercal$, $\begin{bmatrix} -1,\: & 0,\: & 0 \end{bmatrix}^\intercal$, and $\begin{bmatrix} 0,\: & 2,\: & 0 \end{bmatrix}^\intercal$ for the \textit{smallroom} sequence.

\begin{table}[!t]

\begin{center}

{\renewcommand{\arraystretch}{1.1}
\caption{Performance comparison of dense depth estimation in terms of MSE and PSNR. All MSE values are scaled to $\times 10^2$.}
\label{tab:comparison}

\begin{tabular}{c||c|c||c|c}
    \hline
    & \multicolumn{2}{c||}{$classroom$} & \multicolumn{2}{c}{$small room$} \\ \cline{2-5}
    &  MSE$\downarrow$ & PSNR$\uparrow$ &  MSE$\downarrow$ & PSNR$\uparrow$ \\ 
    \hline \hline
    GC-Net~\cite{kendall2017end}           & 0.951 & 20.239 & 5.801 & 12.366  \\ \hline
    PSMNet~\cite{chang2018pyramid}        & 4.127 & 13.844 & 7.862 & 11.045 \\ \hline
    GA-Net~\cite{Zhang_2019_CVPR}         & 2.346 & 16.298 & 4.581 & 13.391  \\ \hline
    360SD-Net~\cite{wang2020360sd}        & 0.218 & 26.625 & 0.581 & 22.361 \\ \hline
    BiFuse~\cite{Wang_2020_CVPR} & 1.803 & 17.481 & 4.108 & 13.866  \\ \hline
    UniFuse~\cite{9353978} & 0.215 & 26.707 & 1.655 & 17.825  \\ \hline
    Proposed  & \multirow{2}{*}{0.205} & \multirow{2}{*}{26.896} & \multirow{2}{*}{0.503} & \multirow{2}{*}{23.018} \\
    (2-cam, known params.) & & & & \\ \hline 
    Proposed  & \multirow{2}{*}{0.259} & \multirow{2}{*}{25.879} & \multirow{2}{*}{0.540} & \multirow{2}{*}{22.706} \\
    (2-cam, unknown params.) & & & & \\ \hline  
    Proposed & \multirow{2}{*}{\textbf{0.198}} & \multirow{2}{*}{\textbf{27.061}} & \multirow{2}{*}{\textbf{0.288}} & \multirow{2}{*}{\textbf{25.415}} \\ 
    (4-cam, known params.) & & & & \\ \hline  
    Proposed & \multirow{3}{*}{0.439} & \multirow{3}{*}{23.587} & \multirow{3}{*}{0.709} & \multirow{3}{*}{21.493} \\ 
    (4-cam, unknown params.,  & & & & \\ 
    without scaling factor) & & & & \\ \hline
    Proposed & \multirow{3}{*}{\textbf{0.193}} & \multirow{3}{*}{\textbf{27.178}} & \multirow{3}{*}{\textbf{0.303}} & \multirow{3}{*}{\textbf{25.193}} \\ 
    (4-cam, unknown params.,  & & & & \\ 
    with scaling factor) & & & & \\ \hline
    \hline
\end{tabular}
}
\end{center}
\end{table}

\begin{figure*}[!t]
    \begin{minipage}{0.08\textwidth} \hspace{1cm} \end{minipage}
    \begin{minipage}{0.22\textwidth}
        \includegraphics[width=1.0\textwidth]{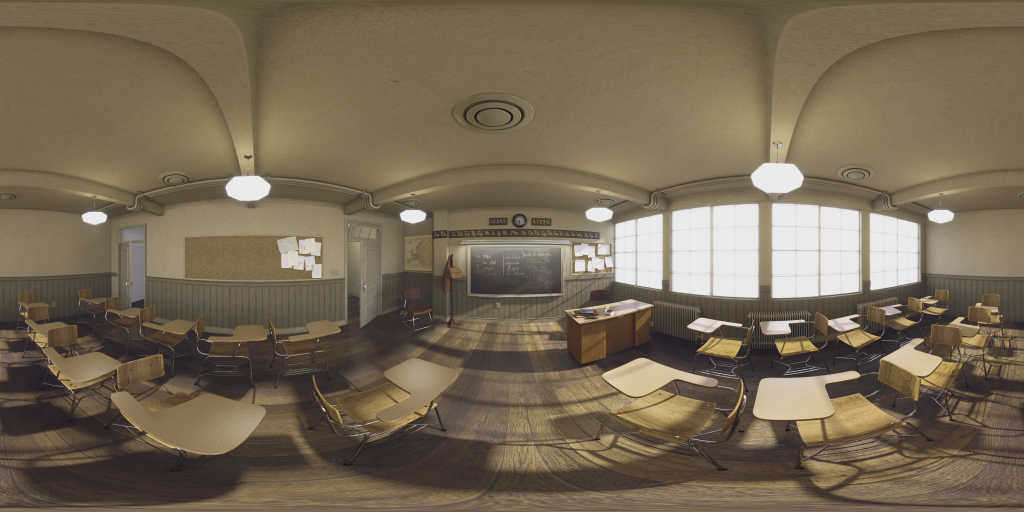} 
        \centerline{(a)} \vspace{0.1cm}
    \end{minipage}
    \begin{minipage}{0.22\textwidth}
        \includegraphics[width=1.0\textwidth]{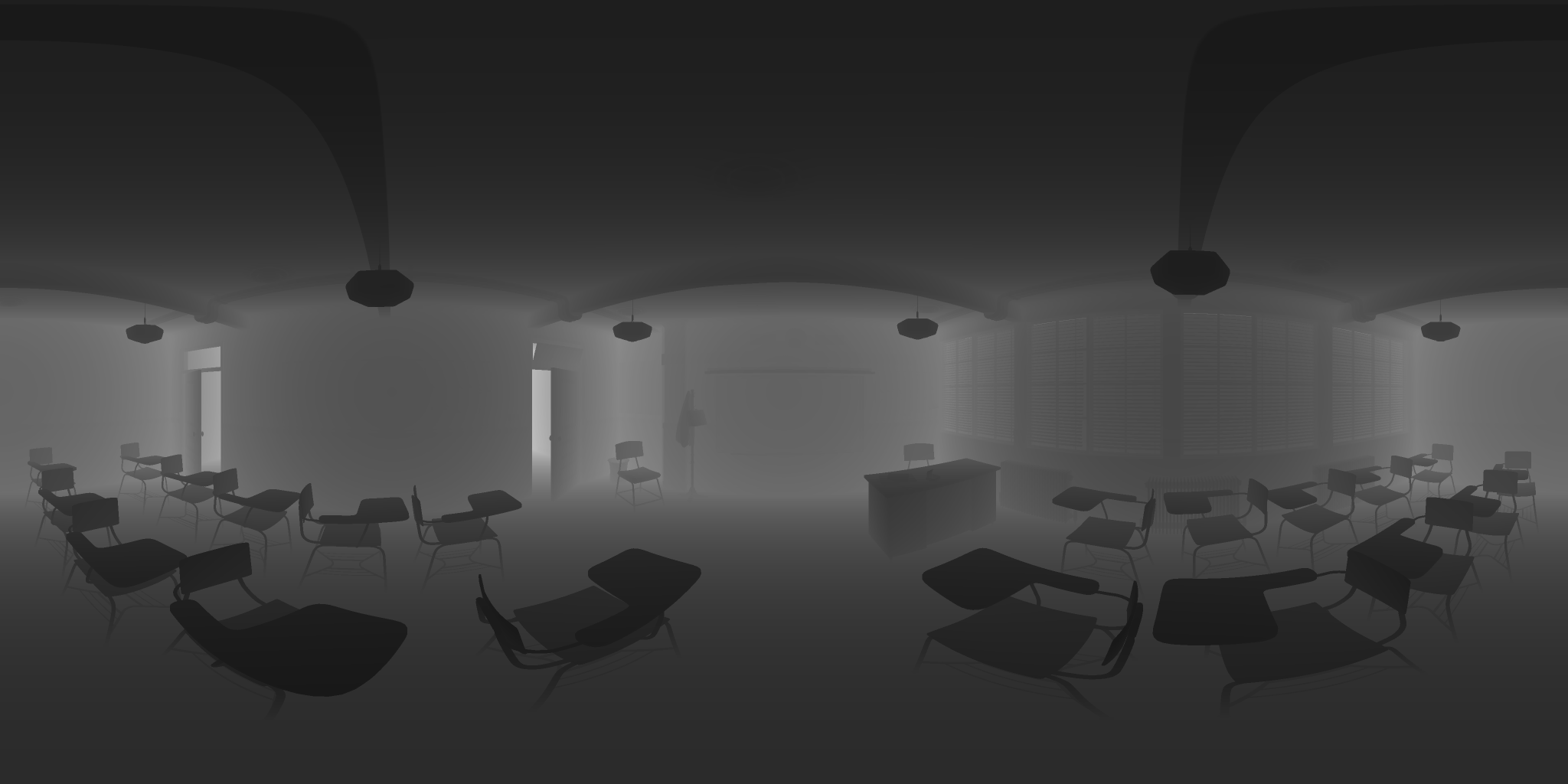} 
        \centerline{(b)} \vspace{0.1cm}
    \end{minipage}
    \begin{minipage}{0.22\textwidth}
        \includegraphics[width=1.0\textwidth]{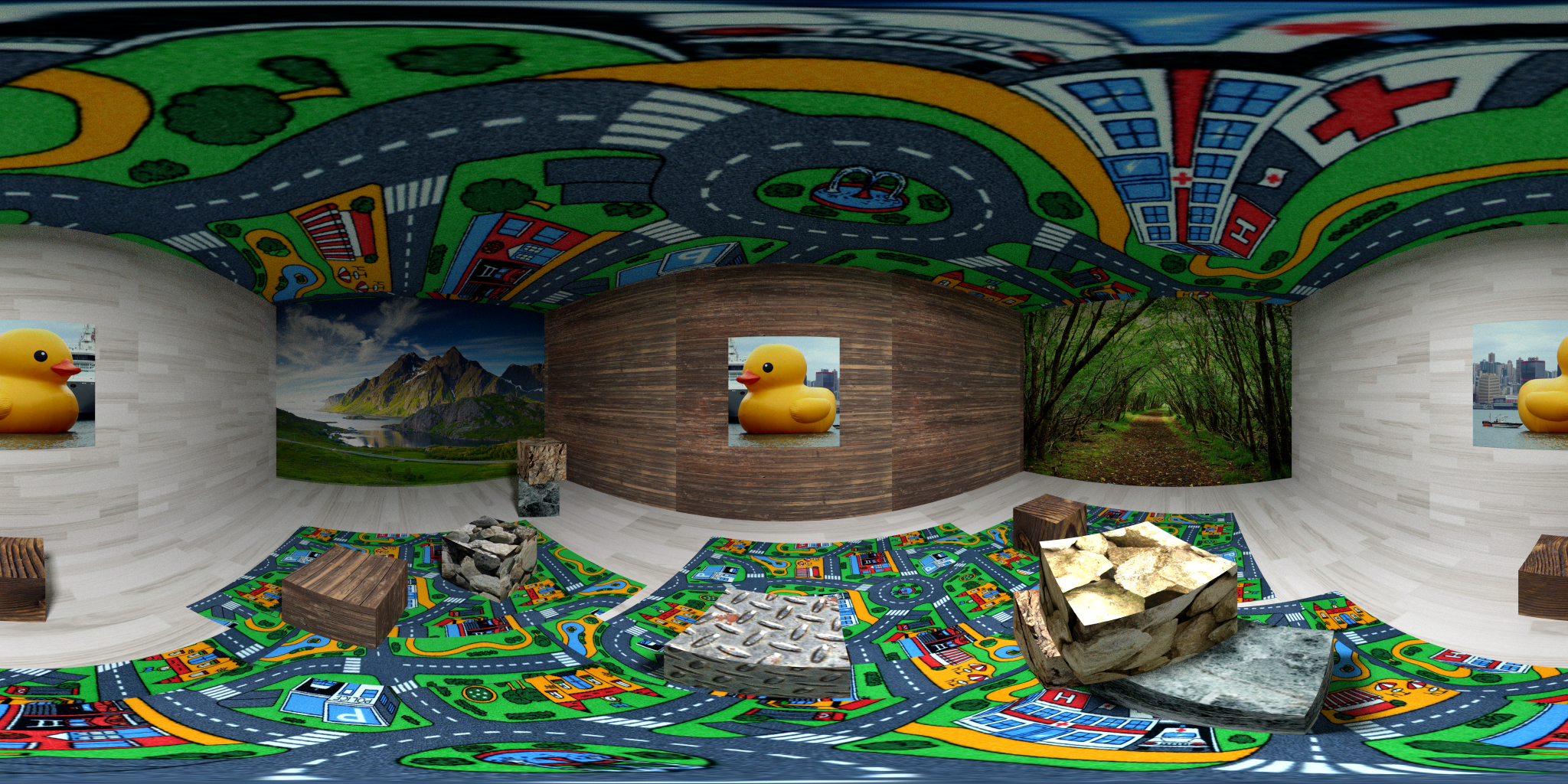} 
        \centerline{(c)} \vspace{0.1cm}
    \end{minipage}
    \begin{minipage}{0.22\textwidth}
        \includegraphics[width=1.0\textwidth]{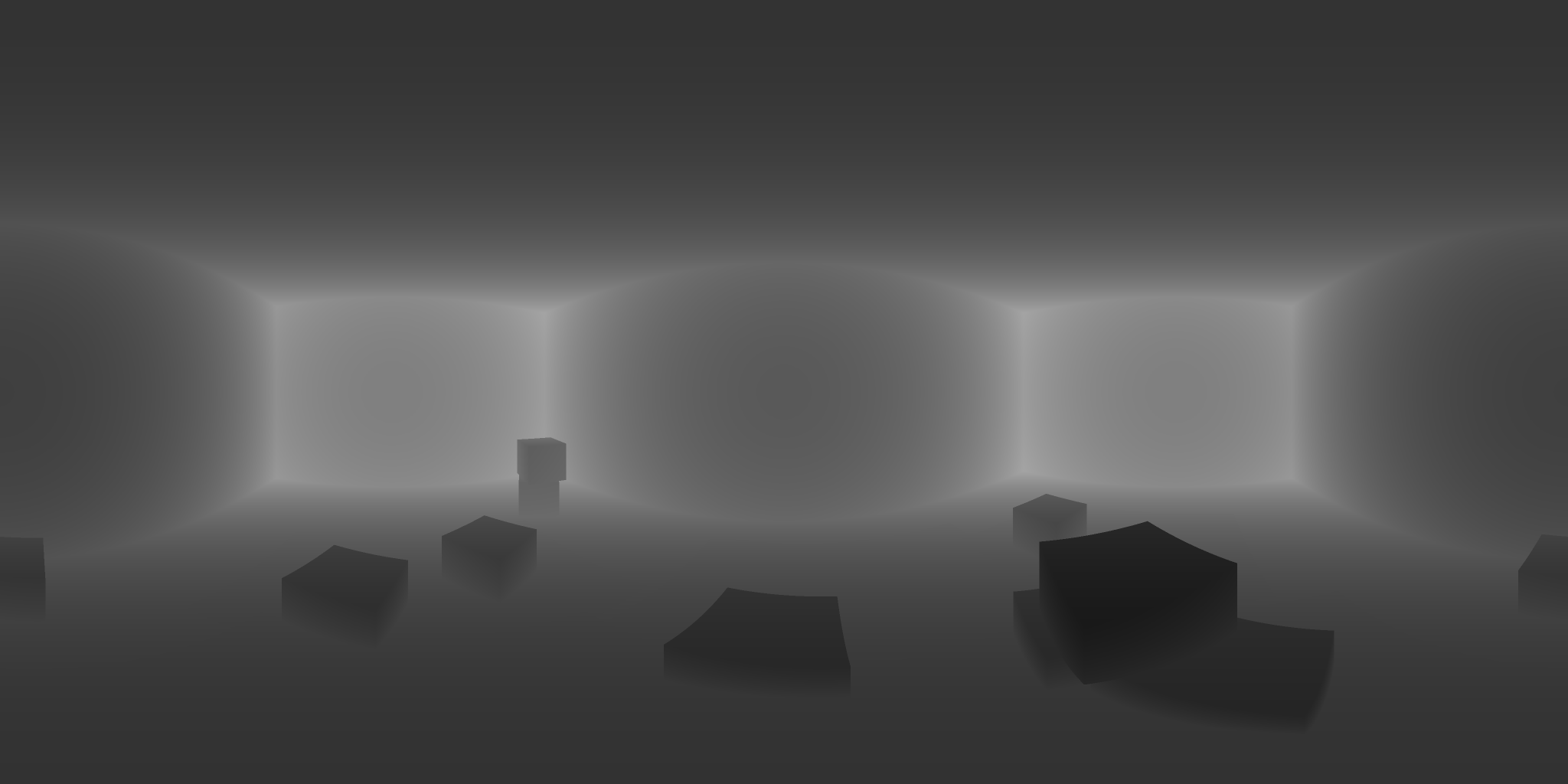} 
        \centerline{(d)} \vspace{0.1cm}
    \end{minipage}
    \\
    \hrule \vspace{0.05cm}
    \begin{minipage}{0.08\textwidth} \centerline{360SD}         \centerline{-Net}\centerline{\cite{wang2020360sd}} \end{minipage}
    \begin{minipage}{0.22\textwidth}
        \includegraphics[width=1.0\textwidth]{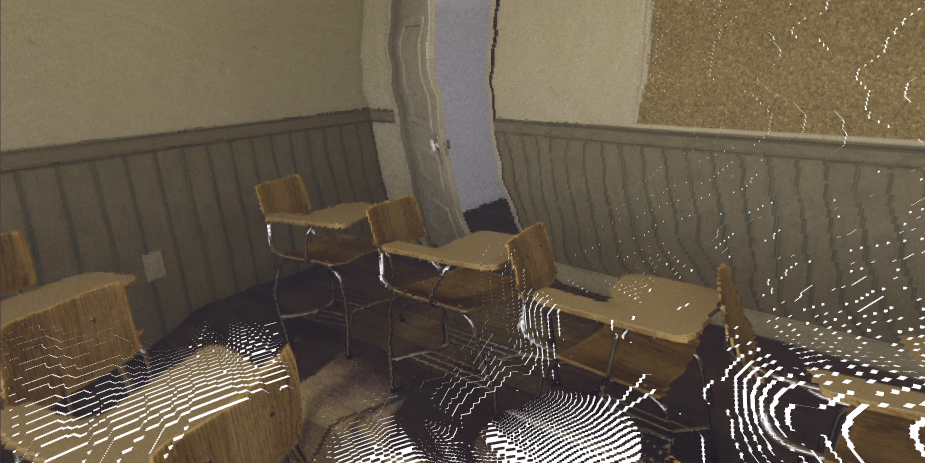} \\  
        \centerline{ } 
    \end{minipage}
    \begin{minipage}{0.22\textwidth}
        \includegraphics[width=1.0\textwidth]{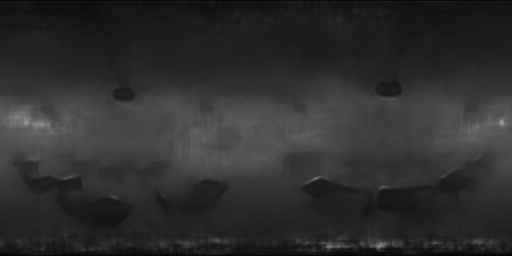} \\ \centerline{{\footnotesize 23.373}}   
    \end{minipage}
    \begin{minipage}{0.22\textwidth}
        \includegraphics[width=1.0\textwidth]{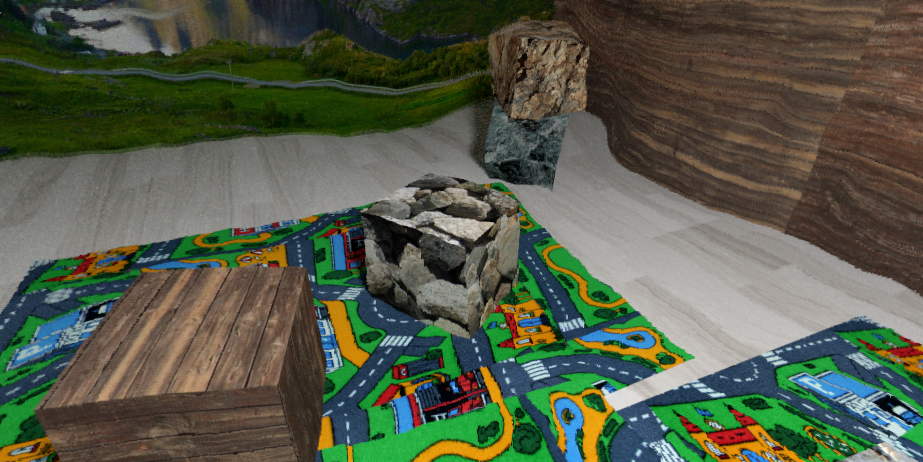}\\  
        \centerline{ }  
    \end{minipage}
    \begin{minipage}{0.22\textwidth}
        \includegraphics[width=1.0\textwidth]{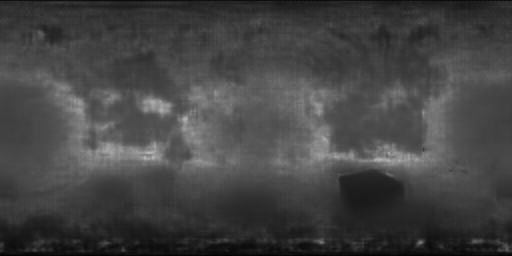} \\  \centerline{{\footnotesize 22.214}}  
    \end{minipage}
    \\
    \begin{minipage}{0.08\textwidth} \centerline{Uni} \centerline{Fuse} \centerline{\cite{9353978}}
    \end{minipage}
    \begin{minipage}{0.22\textwidth}
        \includegraphics[width=1.0\textwidth]{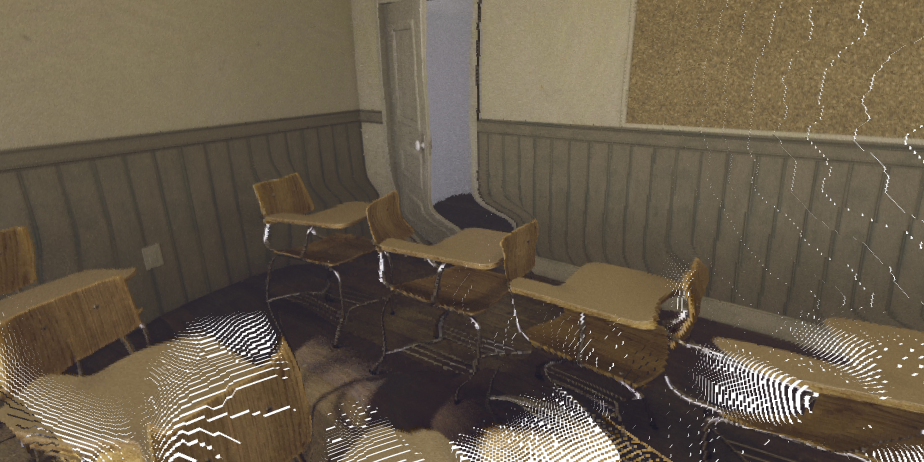} \\  
        \centerline{ } 
    \end{minipage}
    \begin{minipage}{0.22\textwidth}
        \includegraphics[width=1.0\textwidth]{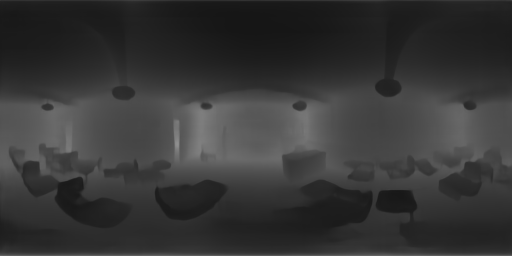} \\  \centerline{{\footnotesize 27.028}}
    \end{minipage}
    \begin{minipage}{0.22\textwidth}
        \includegraphics[width=1.0\textwidth]{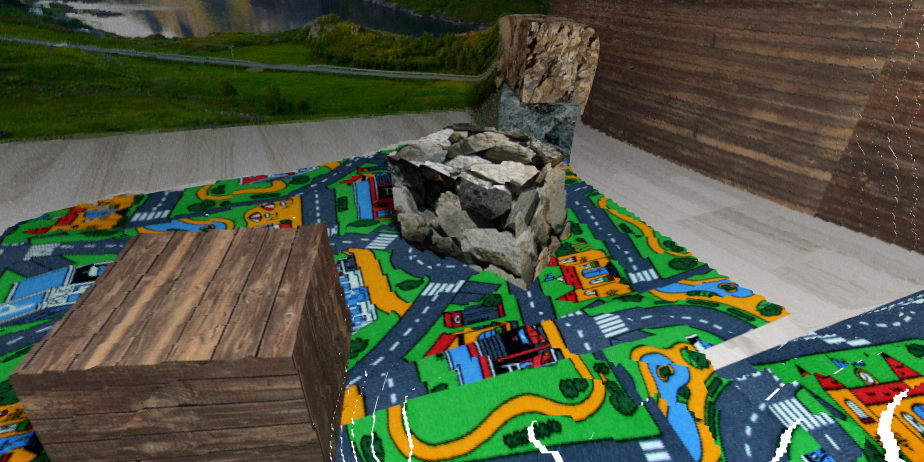} \\  
        \centerline{ } 
    \end{minipage}
    \begin{minipage}{0.22\textwidth}
        \includegraphics[width=1.0\textwidth]{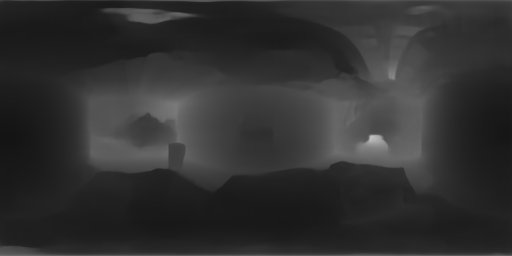} \\  \centerline{{\footnotesize 17.401}}
    \end{minipage}
    \\
    \begin{minipage}{0.08\textwidth} \centerline{Pro} \centerline{-posed} \centerline{} \end{minipage}
    \begin{minipage}{0.22\textwidth}
        \includegraphics[width=1.0\textwidth]{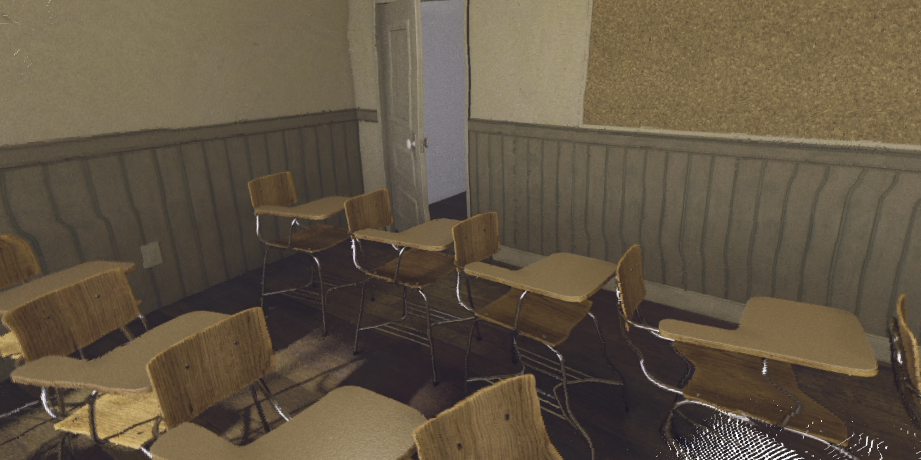} \\ 
        \centerline{ } 
    \end{minipage}
    \begin{minipage}{0.22\textwidth}
        \includegraphics[width=1.0\textwidth]{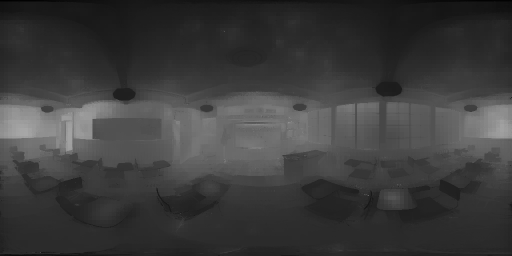}  \\  \centerline{{\footnotesize 27.261}}
    \end{minipage}
    \begin{minipage}{0.22\textwidth}
        \includegraphics[width=1.0\textwidth]{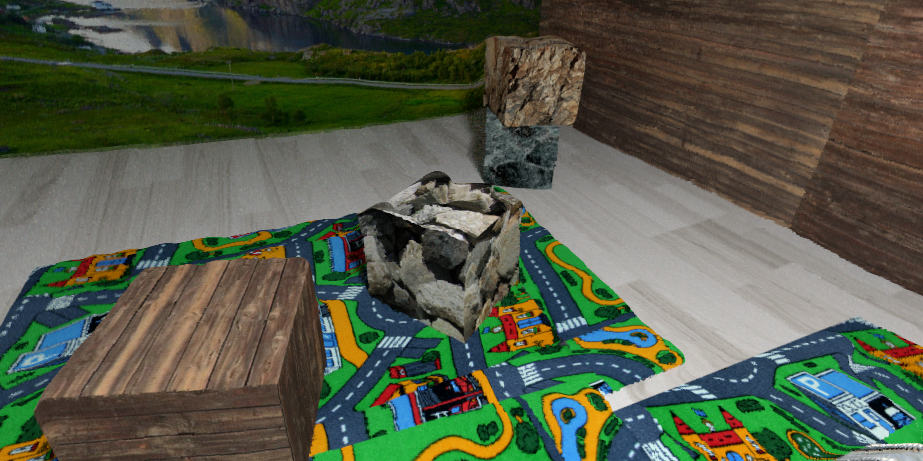}  \\ 
        \centerline{ } 
    \end{minipage}
    \begin{minipage}{0.22\textwidth}
        \includegraphics[width=1.0\textwidth]{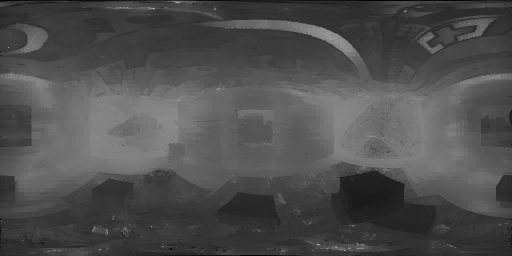}  \\  \centerline{{\footnotesize 24.896}} 
    \end{minipage}
    \\
    \begin{minipage}{0.08\textwidth} \hspace{1.0cm} \end{minipage}
    \begin{minipage}{0.22\textwidth}
        \vspace{0.2cm} \centerline{ (e) \vspace{0.1cm} } 
    \end{minipage}
    \begin{minipage}{0.22\textwidth}
        \vspace{0.2cm} \centerline{ (f) \vspace{0.1cm} } 
    \end{minipage}
    \begin{minipage}{0.22\textwidth}
        \vspace{0.2cm} \centerline{ (g) \vspace{0.1cm}} 
    \end{minipage}
    \begin{minipage}{0.22\textwidth}
        \vspace{0.2cm} \centerline{ (h) \vspace{0.1cm}} 
    \end{minipage}
    \caption{Examples of the estimated dense depth map using the synthesized dataset. (a) Input image. (b) Ground truth depth map. (c) Input Image. (d)  Ground truth depth map. (e) 3D reconstruction using (f). (f) Estimated depth map. (g) 3D reconstruction using (h). (h) Estimated depth map. The numbers below (f) and (h) are the PSNR values of the estimated dense depth map. \vspace{0.3cm}} \label{fig:result_syn}
\end{figure*}

\begin{figure*}[!t]
    \begin{minipage}{0.03\textwidth}
        (a)
    \end{minipage}
    \begin{minipage}{0.31\textwidth}
        \centerline{Ground-truth} \vspace{0.1cm}
        \includegraphics[width=1.0\textwidth]{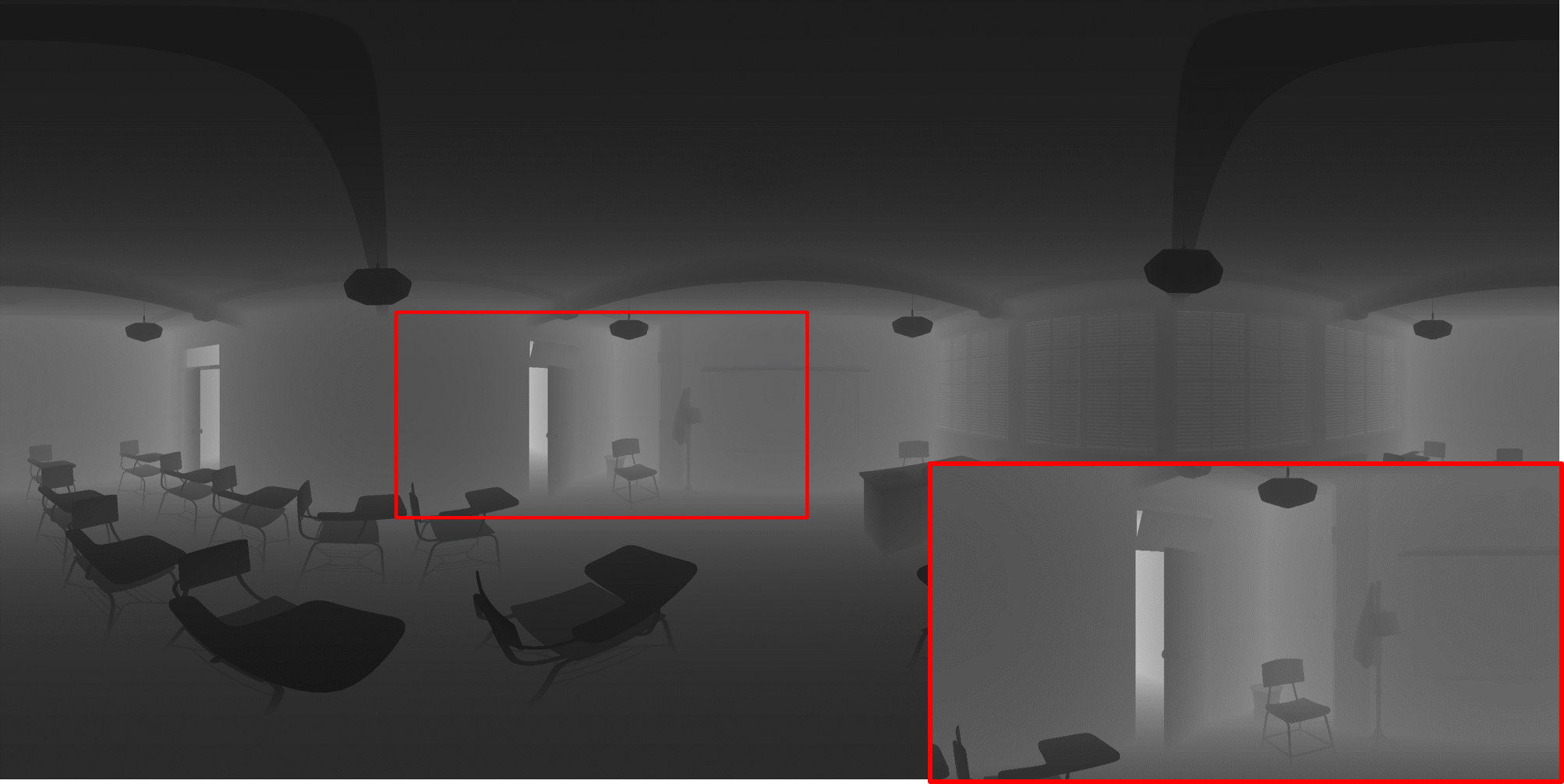} \vspace{0.1cm}
    \end{minipage}
    \begin{minipage}{0.31\textwidth}
        \centerline{With scaling factor} \vspace{0.1cm}
        \includegraphics[width=1.0\textwidth]{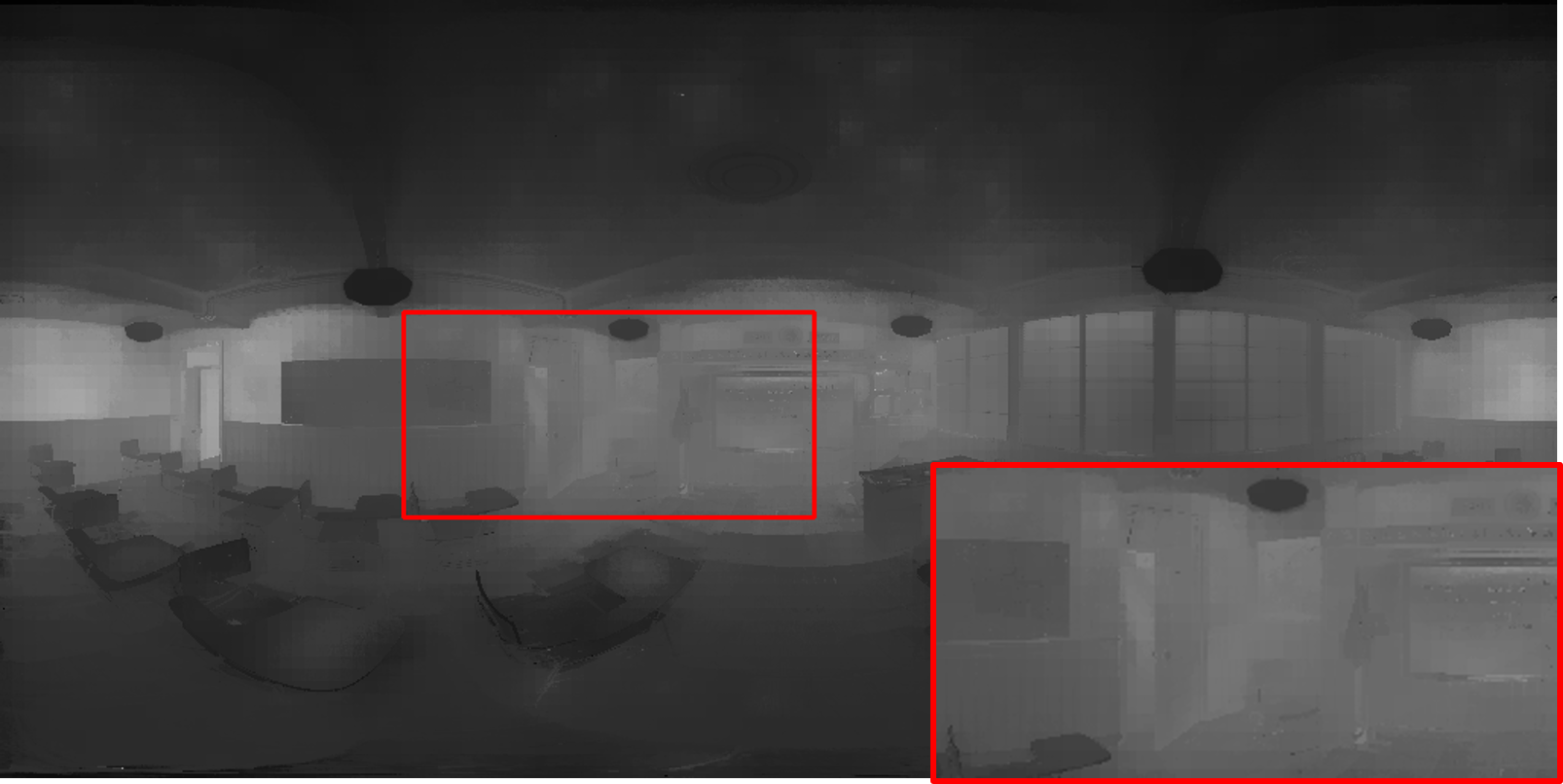}\\ \centerline{ 27.178 }
    \end{minipage}
    \begin{minipage}{0.31\textwidth}
        \centerline{Without scaling factor} \vspace{0.1cm}
        \includegraphics[width=1.0\textwidth]{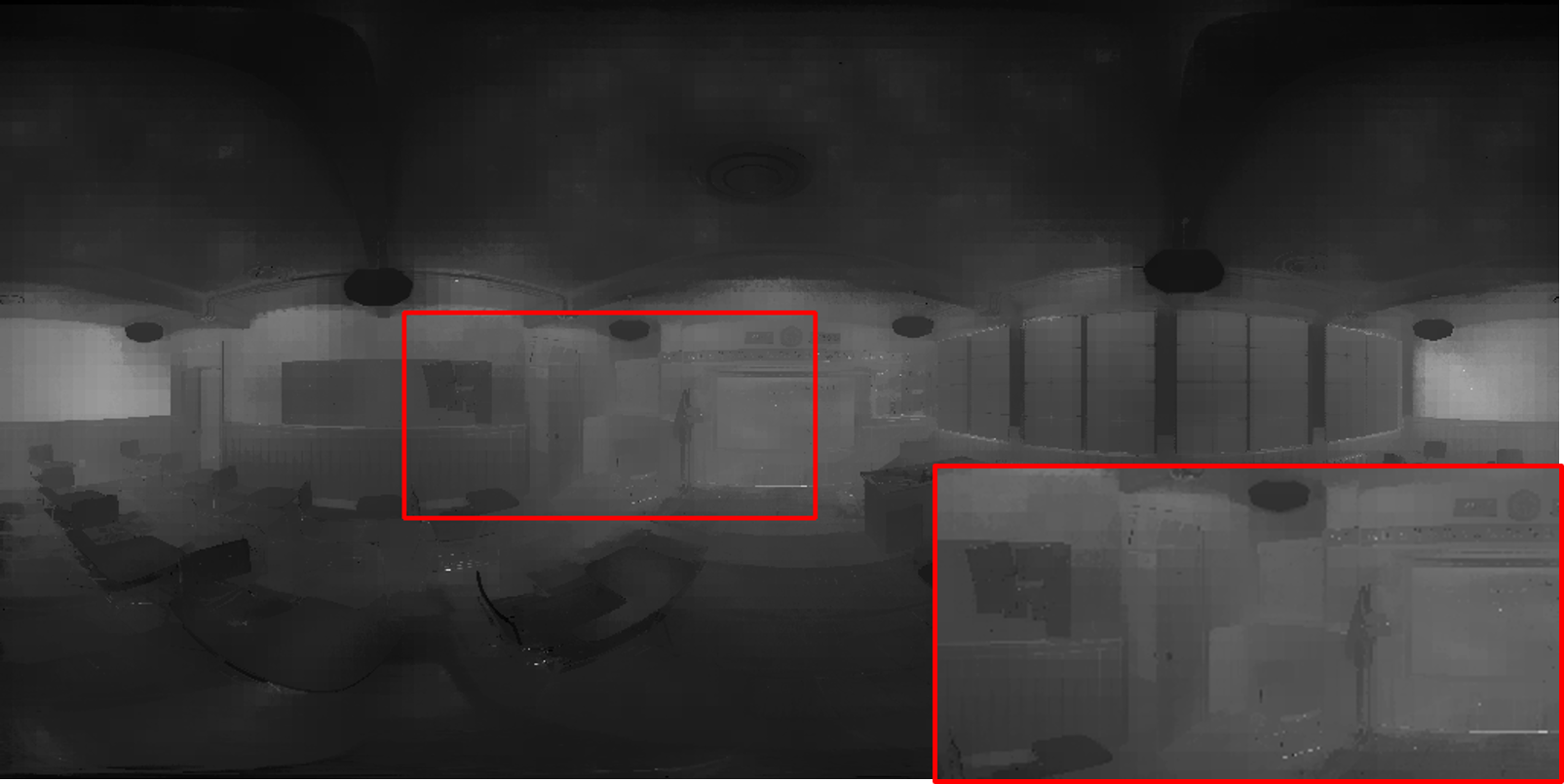} \\ \centerline{ 23.587 }
    \end{minipage} 
    \\
    \begin{minipage}{0.03\textwidth}
        (b)
    \end{minipage}
    \begin{minipage}{0.31\textwidth}
        \includegraphics[width=1.0\textwidth]{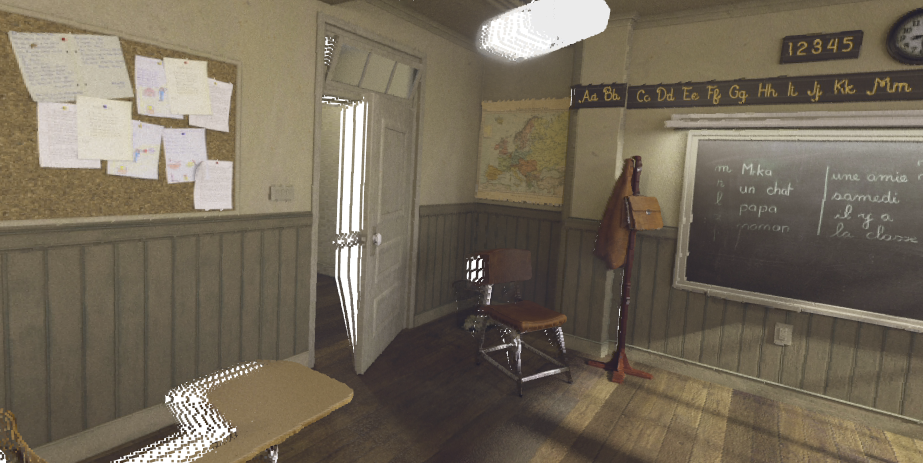} \vspace{0.2cm}
    \end{minipage}
    \begin{minipage}{0.31\textwidth}
        \includegraphics[width=1.0\textwidth]{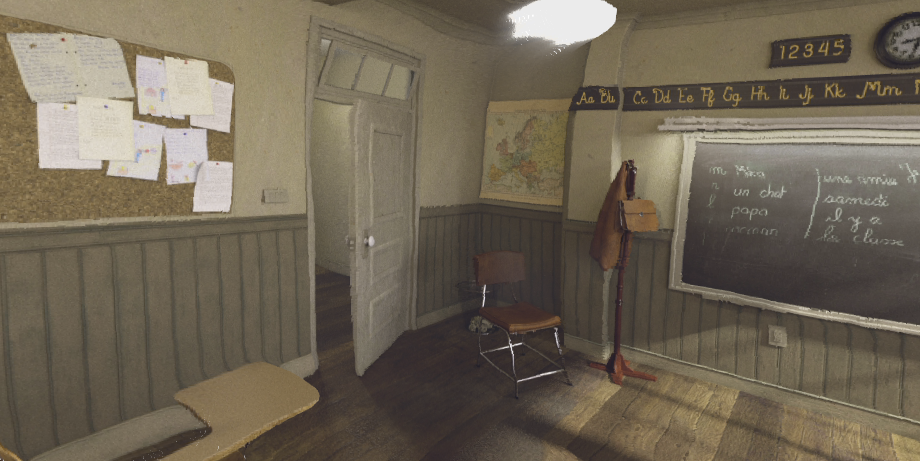} \vspace{0.2cm}
    \end{minipage}
    \begin{minipage}{0.31\textwidth}
        \includegraphics[width=1.0\textwidth]{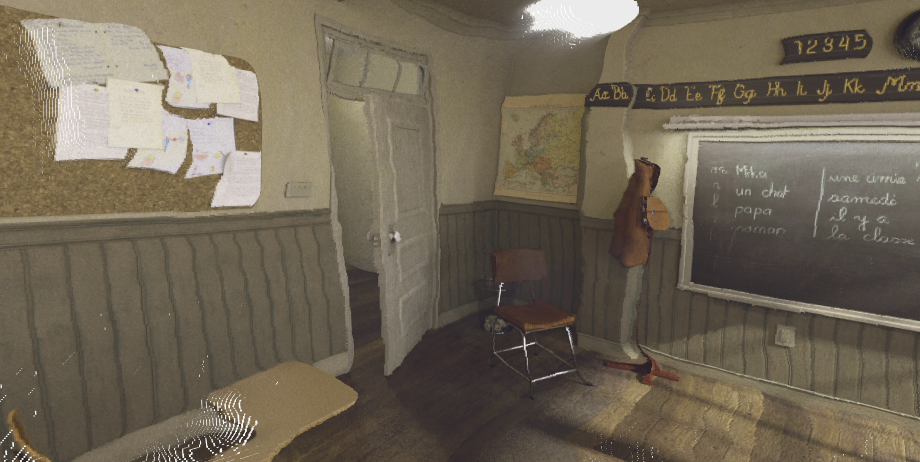} \vspace{0.2cm}
    \end{minipage} 
    \\
    \begin{minipage}{0.03\textwidth}
        (c)
    \end{minipage}
    \begin{minipage}{0.31\textwidth}
        \centering \includegraphics[width=1.0\textwidth]{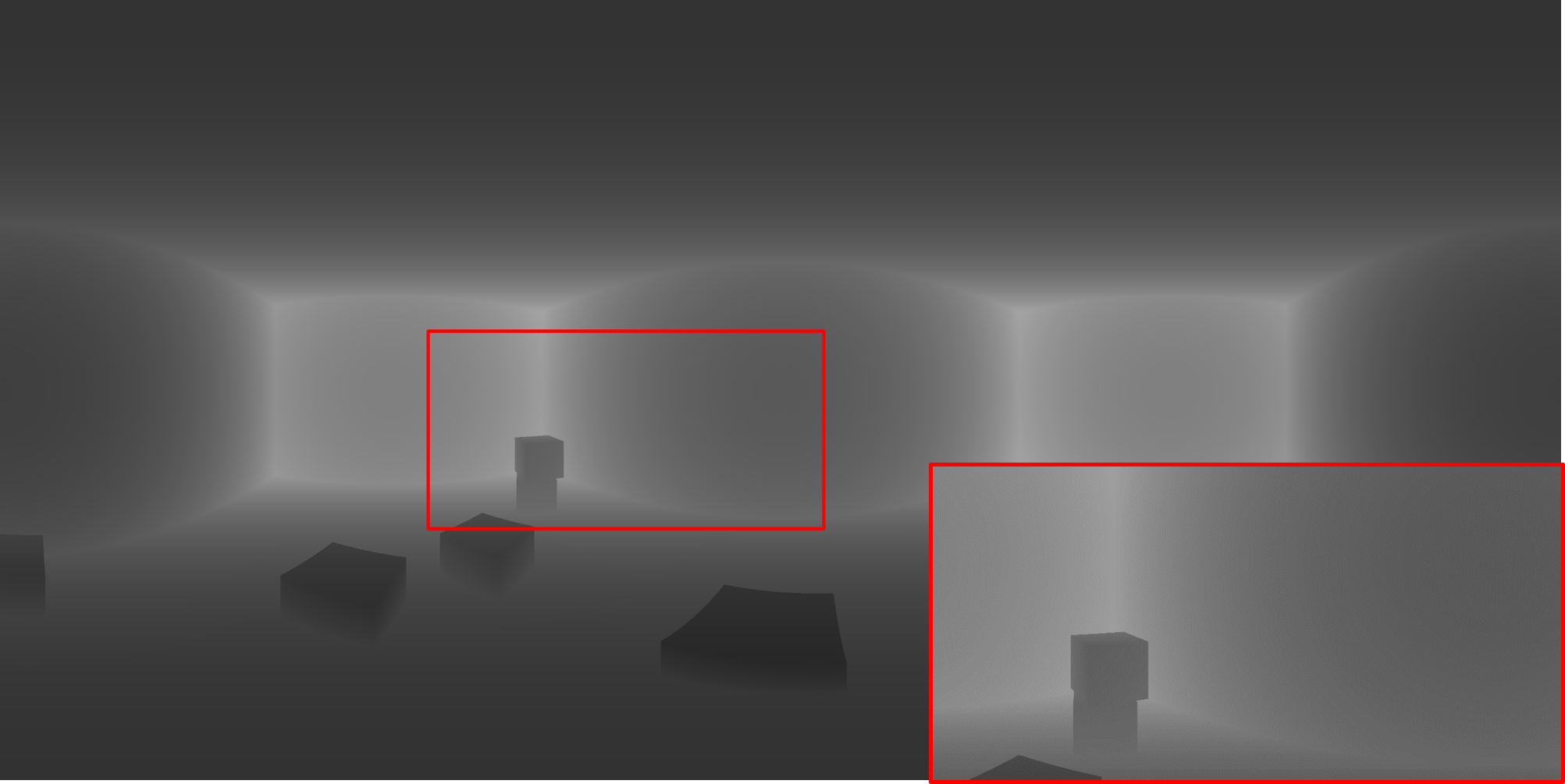} \vspace{0.1cm}
    \end{minipage}
    \begin{minipage}{0.31\textwidth}
        \centering \includegraphics[width=1.0\textwidth]{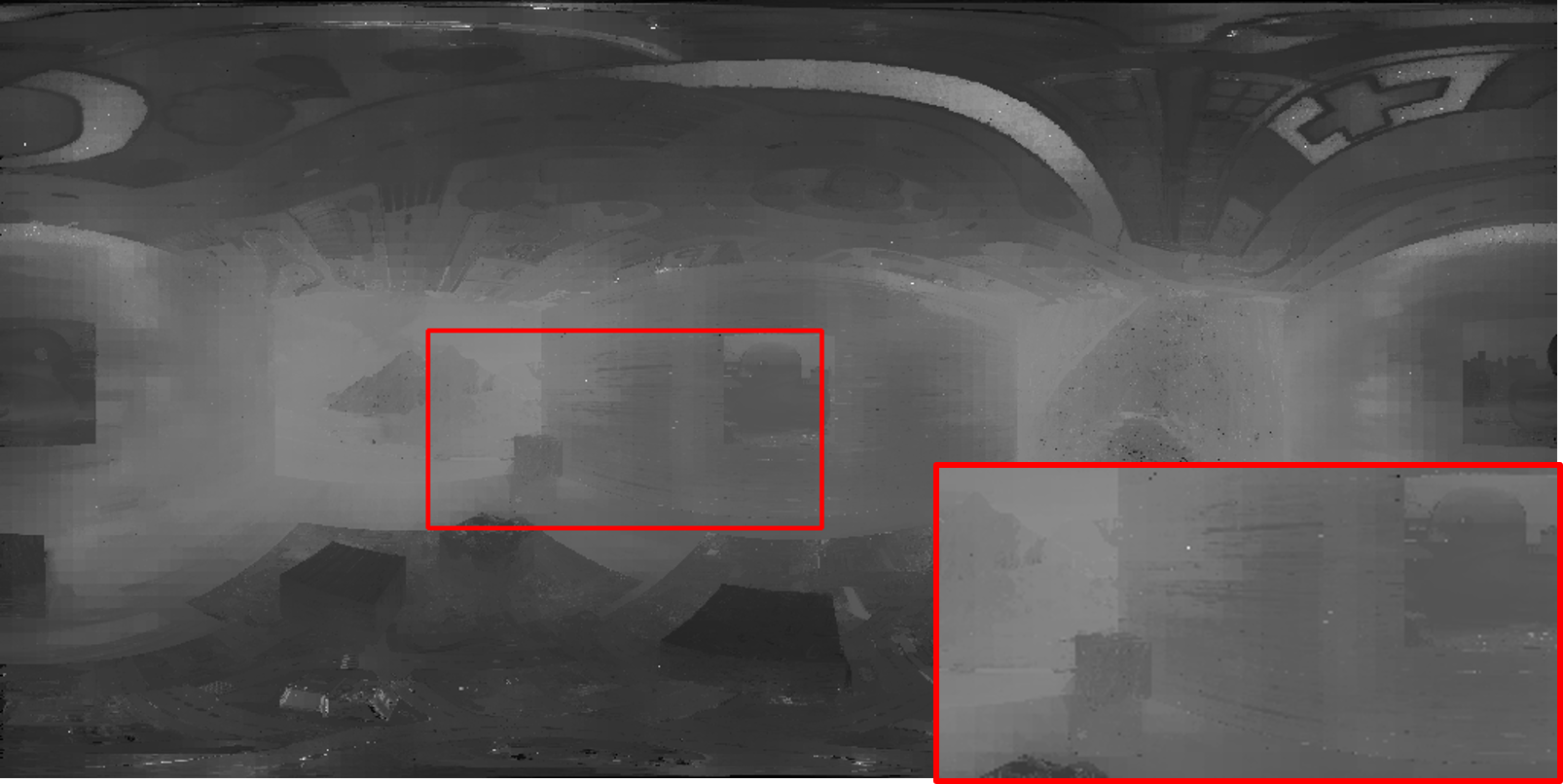}\\ \centerline{ 25.193 }
    \end{minipage}
    \begin{minipage}{0.31\textwidth}
        \centering \includegraphics[width=1.0\textwidth]{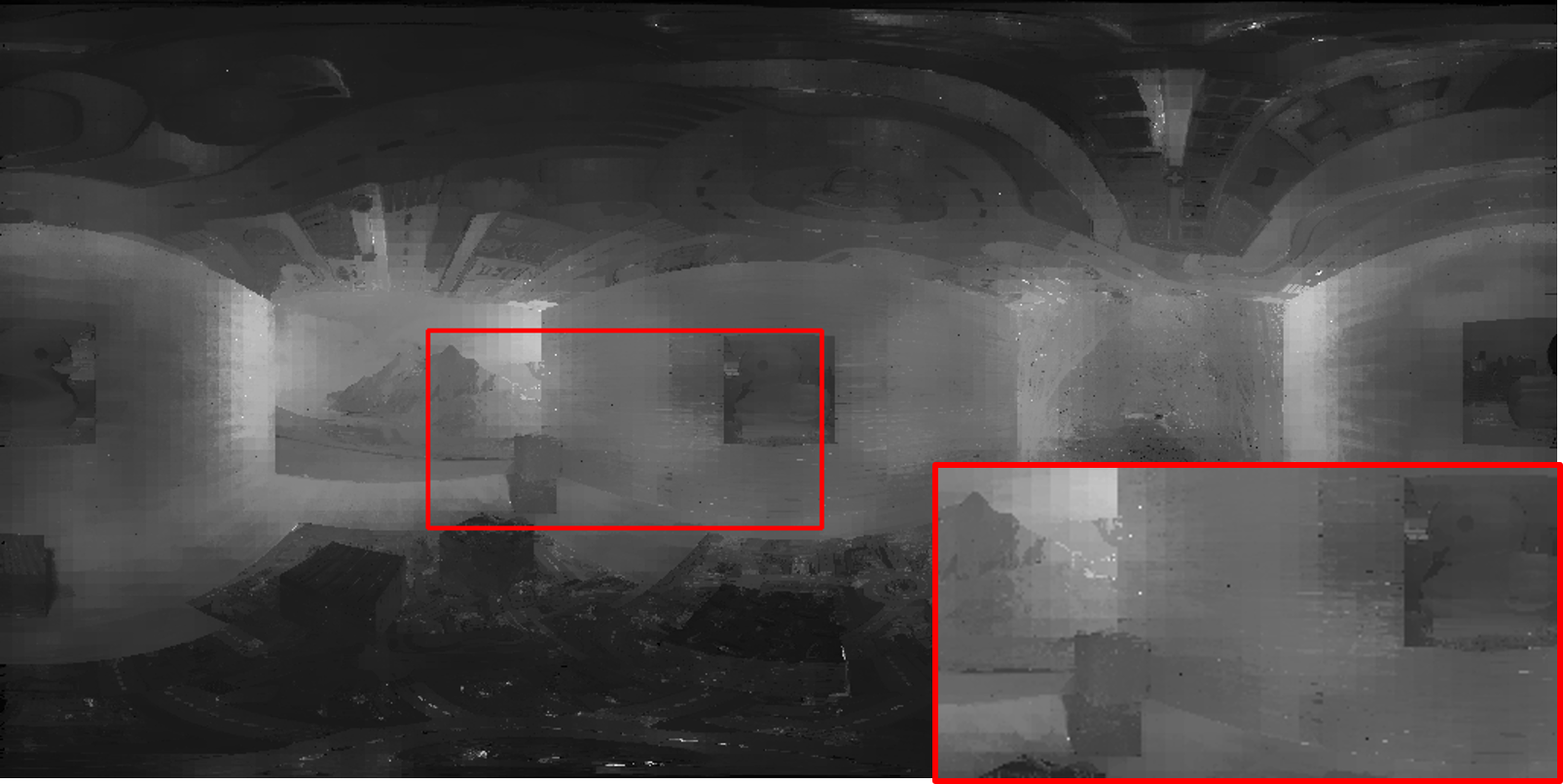}\\ \centerline{ 21.493 }
    \end{minipage} 
    \\
    \begin{minipage}{0.03\textwidth}
        (d)
    \end{minipage}
    \begin{minipage}{0.31\textwidth}
        \includegraphics[width=1.0\textwidth]{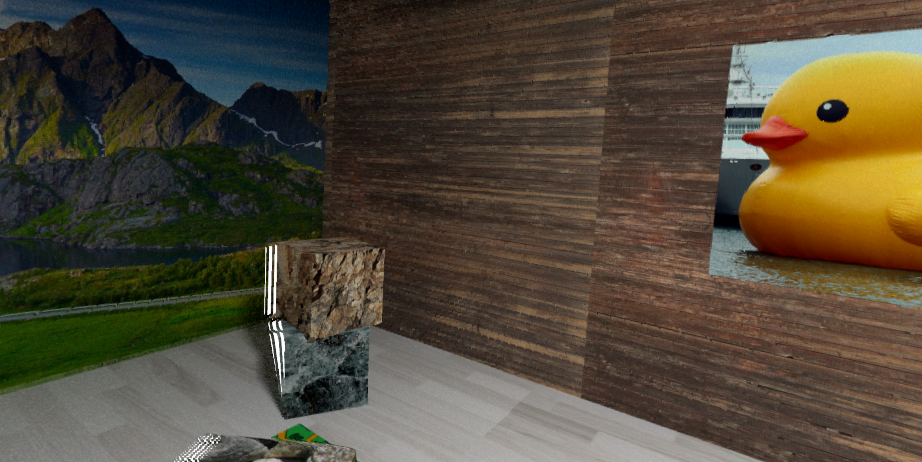} \vspace{0.1cm}
    \end{minipage}
    \begin{minipage}{0.31\textwidth}
        \includegraphics[width=1.0\textwidth]{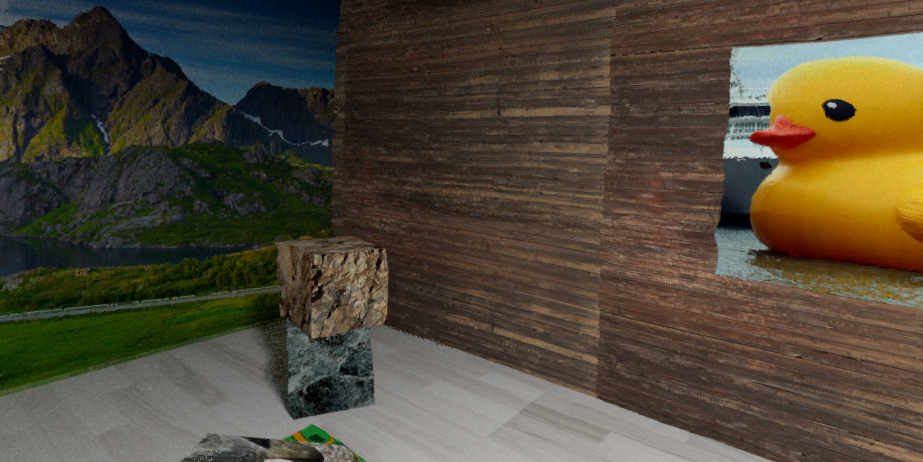} \vspace{0.1cm} 
    \end{minipage}
    \begin{minipage}{0.31\textwidth}
        \includegraphics[width=1.0\textwidth]{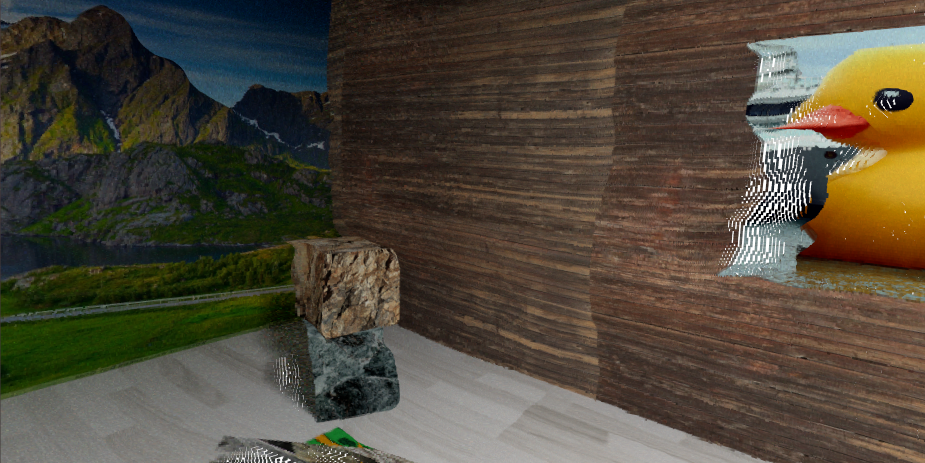} \vspace{0.1cm}
    \end{minipage} 
    \caption{Visual comparisons of the estimated dense depth map applying/not applying the scaling factor in \eref{eq:scaling}. (a) Dense depth maps with the zoomed and cropped. (b) 3D reconstruction using (a). (c) Dense depth maps with the zoomed and cropped. (d) 3D reconstruction using (c). The numbers below the estimated dense depth maps are PSNR.}\label{fig:nscale}
\end{figure*}

The estimation performance is tabulated in \tref{tab:comparison} in terms of mean square error~(MSE) and peak signal-to-noise ratio~(PSNR). And the example images of the current state-of-art algorithms and the proposed pipeline with MSE and PSNR are shown in \fref{fig:result_syn}. The 3D views of the resulting point cloud using Open3D~\cite{Zhou2018} are in the examples. Overall, the estimation performance was slightly improved for the proposed algorithm with two views. By multiple camera settings with the scaling coefficients in \eref{eq:t_vector}, the estimation performance was substantially improved relative to the baseline and other methods. 
Although GC-Net~\cite{kendall2017end}, PSMNet~\cite{chang2018pyramid}, and GA-Net~\cite{Zhang_2019_CVPR} become commonplace in disparity estimation using a deep neural network framework, the algorithms were developed for perspective images. This implies that the models could not represent the characteristic of spherical images exactly, as demonstrated by their relative lower depth estimation accuracy in \tref{tab:comparison}. The algorithms developed for 360\degree\: images, 360SD-Net~\cite{wang2020360sd} and UniFuse~\cite{9353978}, produced decent results. However, the algorithms require additional inputs rather than spherical images, such as a polar angle map for the 360SD-Net algorithm and cubemap projections for the UniFuse and BiFuse algorithms. Since the UniFuse and BiFuse~\cite{Wang_2020_CVPR} algorithms take a monocular spherical image as an input, their performances were not consistent over test images. For 360SD-Net, the algorithm only works for the limited camera configuration aligned with a vertical displacement between camera views. Furthermore, it requires intensive memory resources and computational cost due to its 3D convolutional structure. On the other hand, as proved in \tref{tab:comparison}, the proposed framework can estimate depth consistently and accurately without any restriction of camera configuration.

In order to prove the effectiveness of the proposed scaling factor, we also compared the examples of the estimated depth applying/not applying it in \fref{fig:nscale}.  As demonstrated in the highlighted regions in the red boxes and the higher values of PSNR, the proposed scaling factor improved the accuracy of depth estimation, which also helped reconstruct better quality 3D images.
Besides, the results in \tref{tab:comparison}, \fref{fig:result_syn}, and \fref{fig:nscale} demonstrate that dense depth estimation accuracy can be improved by jointly finding solutions from delicately aligned multiview images. 

\begin{table}[!t]

\begin{center}

{\renewcommand{\arraystretch}{1.1}
\caption{The analysis of estimation performance with different window sizes and different numbers of cameras with unknown $\mathbf{R}$ and $\mathbf{t}$ parameters. All MSE values are scaled to $\times 10^2$.}
\label{tab: win&cam}

\begin{tabular}{c|c||c|c||c|c}
    \hline
    \multicolumn{2}{c||}{} & \multicolumn{2}{c||}{$classroom$} & \multicolumn{2}{c}{$smallroom$} \\
    \hline\hline
    Camera & Window & MSE$\downarrow$ & PSNR$\uparrow$ & MSE$\downarrow$ & PSNR$\uparrow$ \\
    \hline \hline
    2 & 5 & 0.341 & 24.694 & 0.606 & 22.181 \\
    2 & 7 & \textbf{0.259} & \textbf{25.879} & \textbf{0.540} & \textbf{22.706} \\
    2 & 9 & 0.276 & 25.609 & 0.748 & 21.268 \\ \hline
    3 & 5 & 0.296 & 25.289 & 0.589 & 22.311 \\
    3 & 7 & \textbf{0.220} & \textbf{26.588} & \textbf{0.449} & \textbf{23.508} \\
    3 & 9 & 0.264 & 25.805 & 0.571 & 22.448 \\ \hline
    4 & 5 & 0.252 & 26.022 & 0.427 & 23.728 \\
    4 & 7 & \textbf{0.193} & \textbf{27.178} & \textbf{0.303} & \textbf{25.193} \\
    4 & 9 & 0.309 & 25.109 & 0.358 & 24.475 \\ \hline
\end{tabular}
}
\end{center}
\end{table}

\begin{figure}[!t]
    \centering
    \captionsetup{justification=centering}
    \includegraphics[scale=0.6]{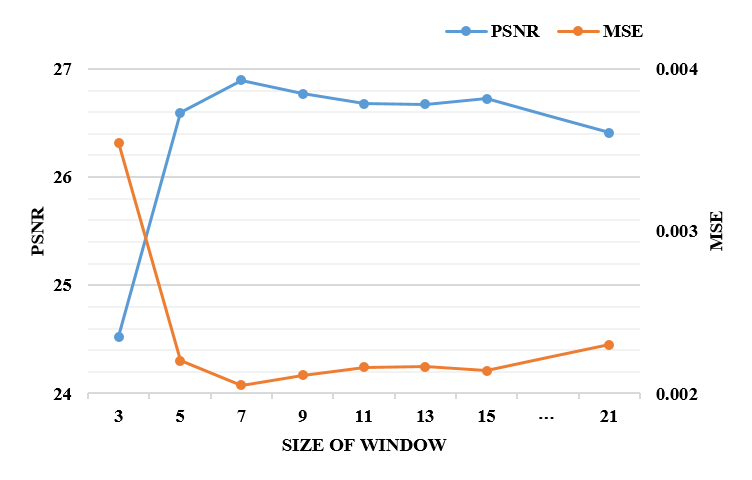}
    \caption{Comparison of dense depth estimation performance using different window sizes for $classroom$ dataset. In this experiment, dense depth maps were estimated using the known $\mathbf{R}$ and $\mathbf{t}$ parameters to examine the effect of window size on the estimation accuracy only. The best performance appears in window size $7 \times 7$.}
    \label{fig: window}
\end{figure}

The dense depth estimation performance was measured using different numbers of cameras and different window sizes $\Omega$ to analyze the effectiveness of virtual depth. The analysis results are summarized in \tref{tab: win&cam}. The estimation accuracy tended to improve as the window size increased because too-small windows could generate noisy depth maps. On the contrary, the estimation accuracy improvement stopped at a certain point. This is because a too-large window size had a higher chance to mismatch the regions where the depth was discontinuous and generate smoother depth maps with fewer details. This analysis also proved in \fref{fig: window} that examined the effect of window size on the accuracy performance for the \textit{classroom} sequence. Based on the analysis, we chose the window size as $7 \times 7$ in the experiments of the \textit{classroom} and \textit{smallroom} sequences.

\begin{figure*}[!t]
    \centering
    \begin{tabular}{@{}c@{~}c@{~}c@{}}
    \includegraphics[width=0.32\textwidth]{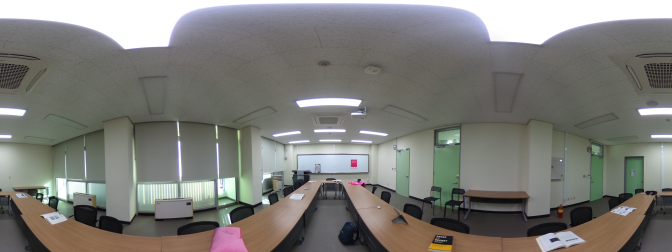} & 
    \includegraphics[width=0.32\textwidth]{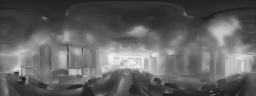} & 
    \includegraphics[width=0.32\textwidth]{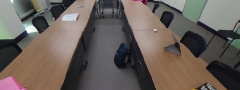} \\
    \includegraphics[width=0.32\textwidth]{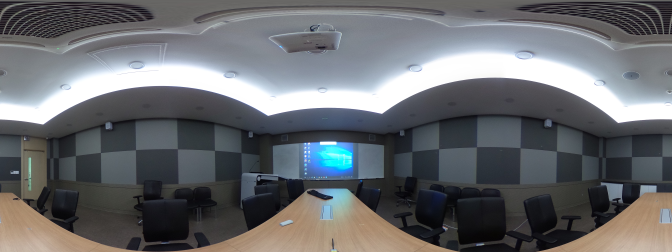} & 
    \includegraphics[width=0.32\textwidth]{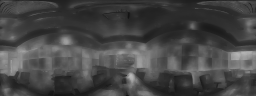} & 
    \includegraphics[width=0.32\textwidth]{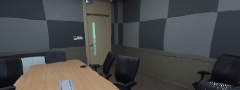} \\
    \includegraphics[width=0.325\textwidth]{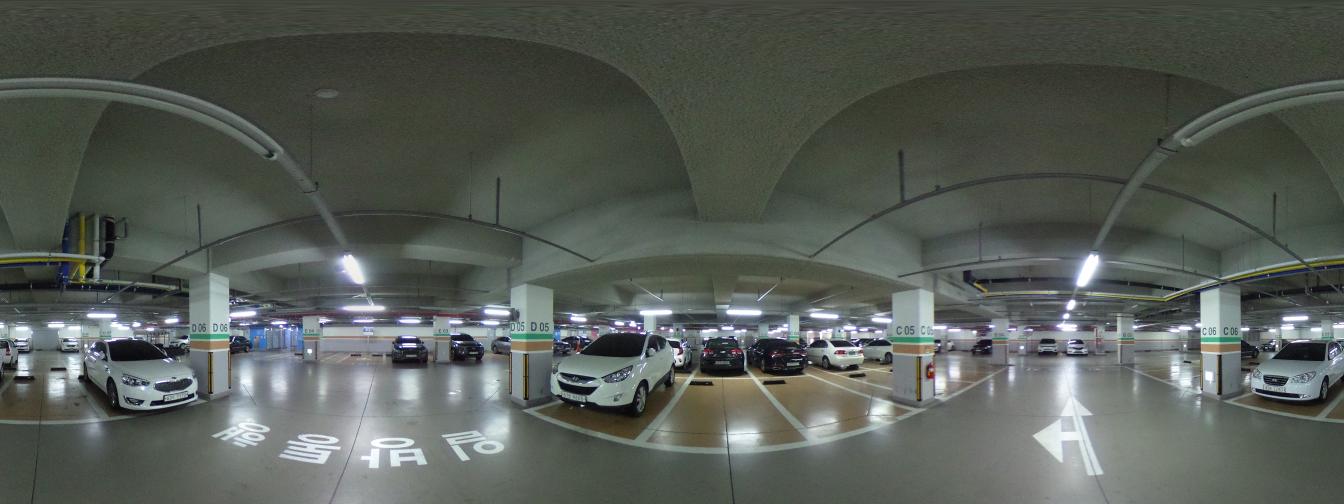} & 
    \includegraphics[width=0.325\textwidth]{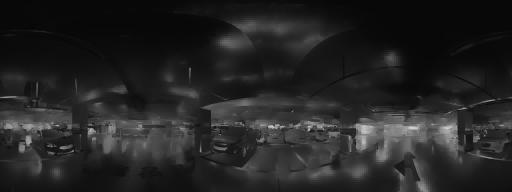} & 
    \includegraphics[width=0.325\textwidth]{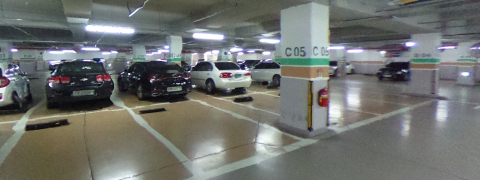} \\
    \includegraphics[width=0.32\textwidth]{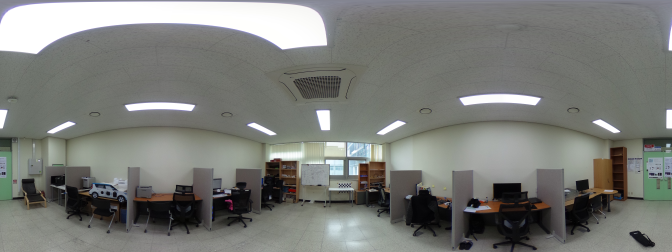} & 
    \includegraphics[width=0.32\textwidth]{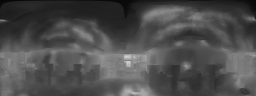} & 
    \includegraphics[width=0.32\textwidth]{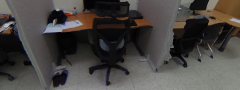} \\
    (a) & (b) & (c) \\
    \end{tabular}
    \caption{Examples of the estimated dense depth map using the proposed pipeline. (a) Input image. (b) Estimated depth map. (c) 3D Reconstruction. From top to bottom, there are 9, 5, 5, and 10-view images in the test image sets. The resultant dense depth maps were estimated using the multiview images.} \label{fig:result_real}
\end{figure*}

\begin{figure*}[!t]
    \centering
    \begin{tabular}{@{}c@{~}c|c@{~}c@{}}
    \includegraphics[width=0.235\textwidth]{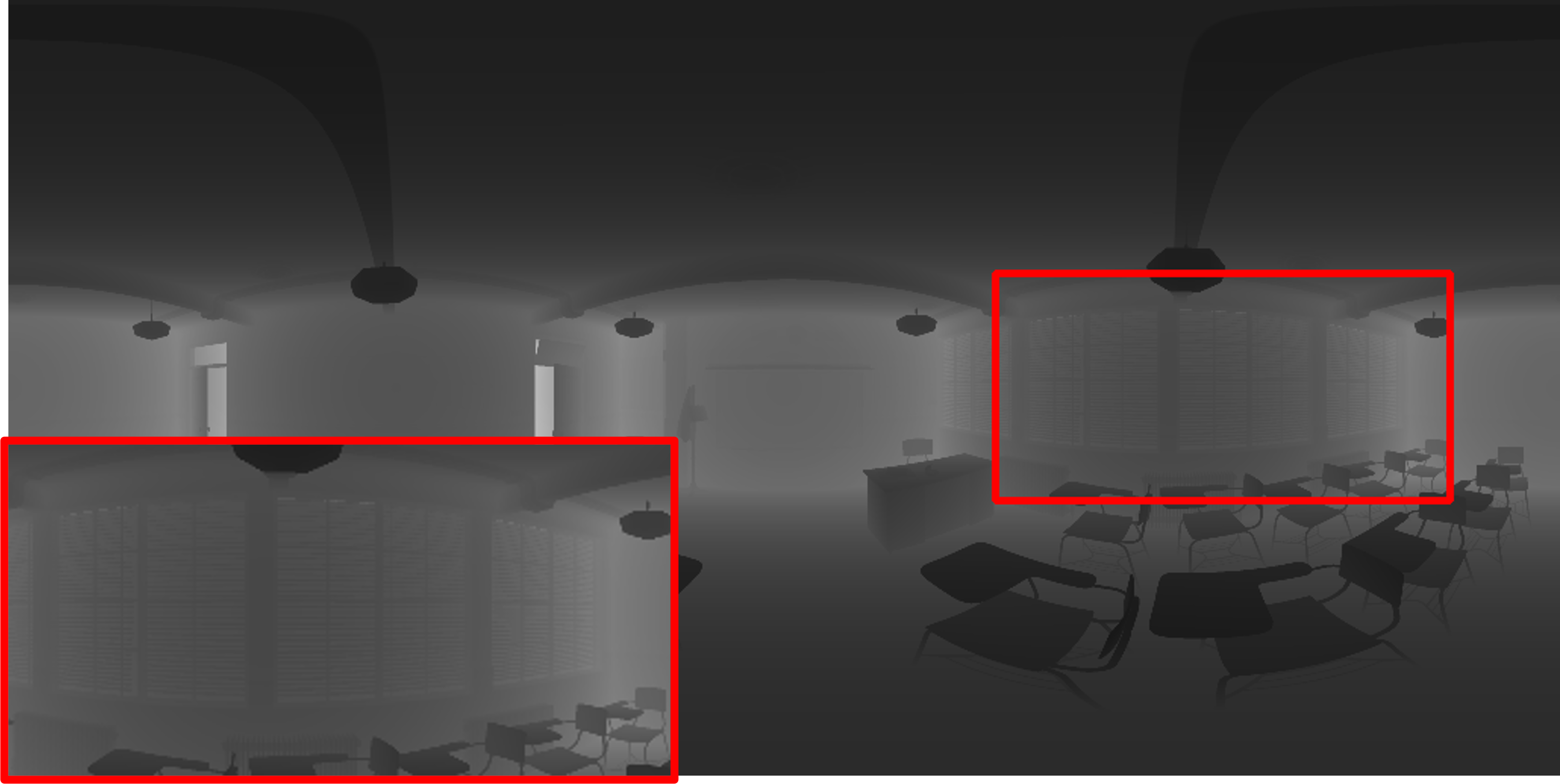} & 
    \includegraphics[width=0.235\textwidth]{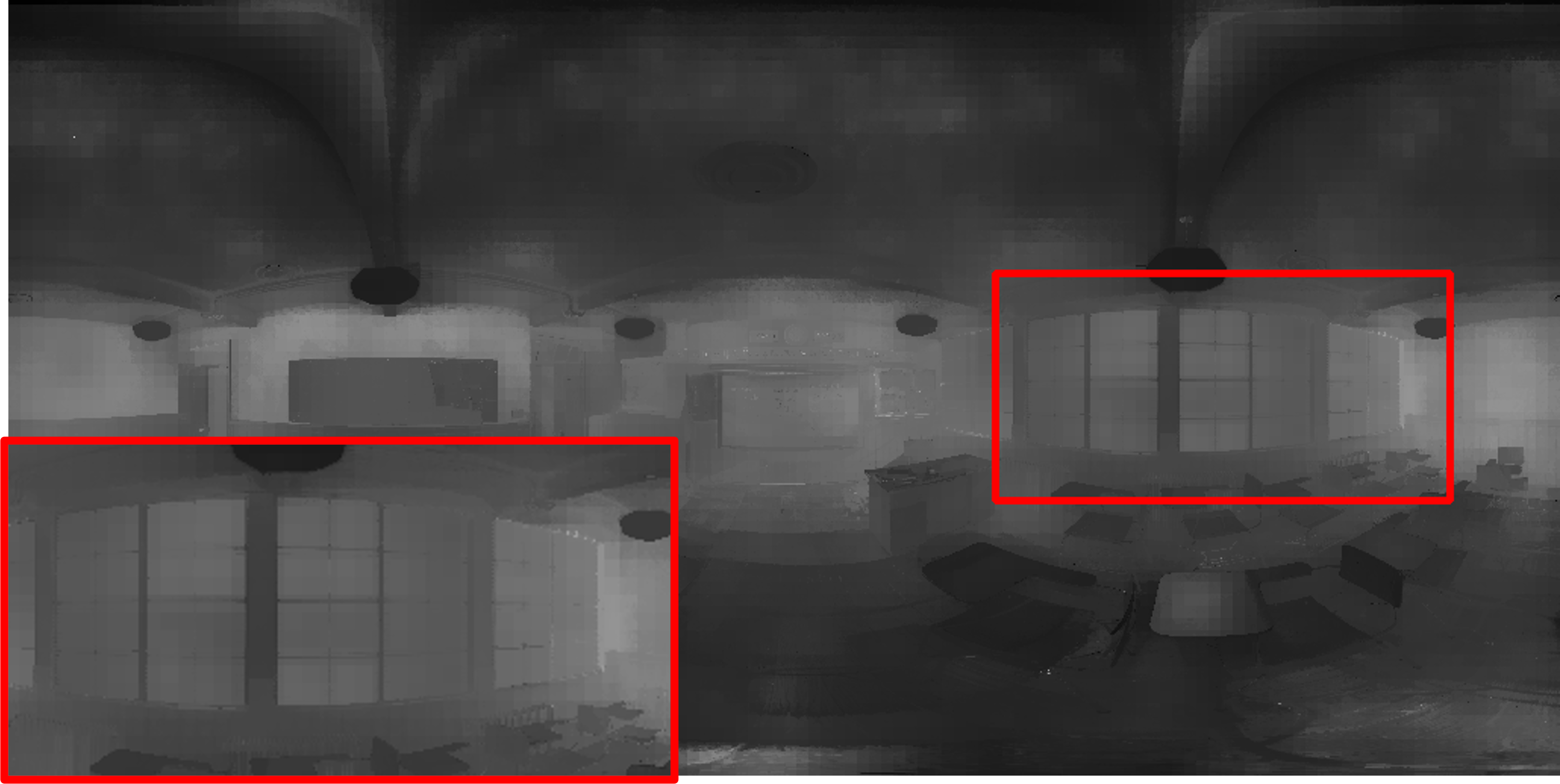} & 
    \includegraphics[width=0.235\textwidth]{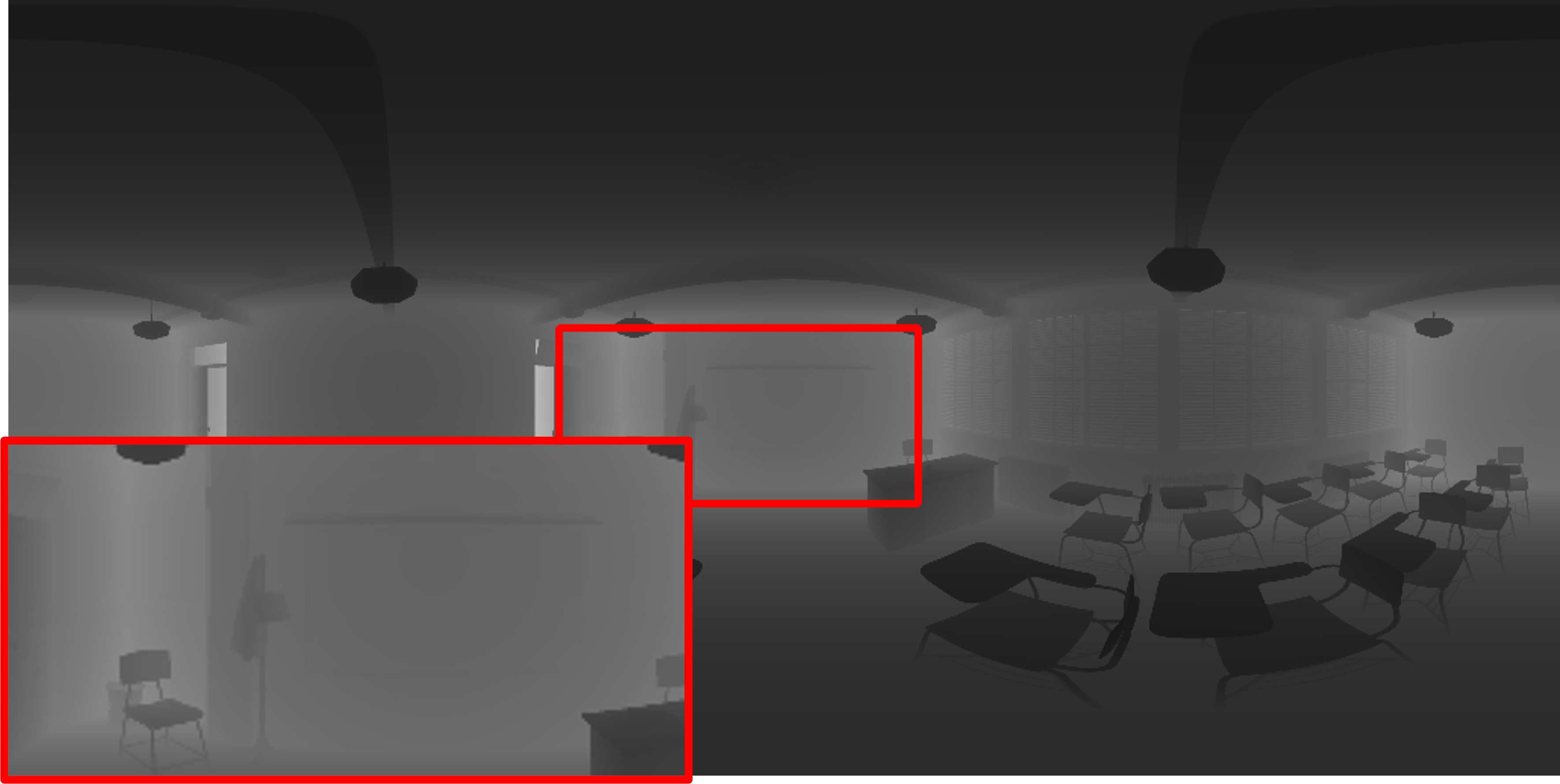} & 
    \includegraphics[width=0.235\textwidth]{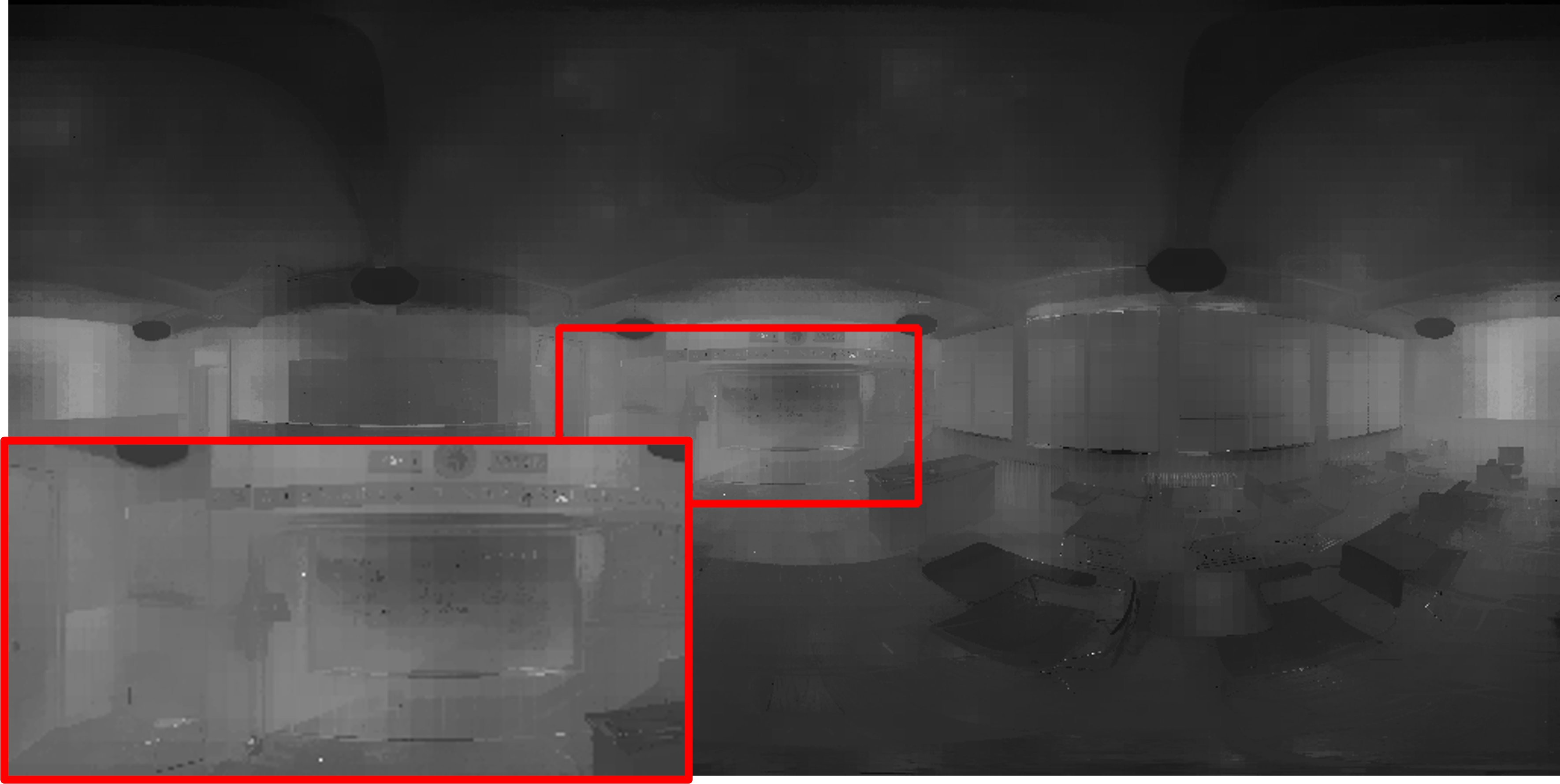}
    \\
    \includegraphics[width=0.235\textwidth]{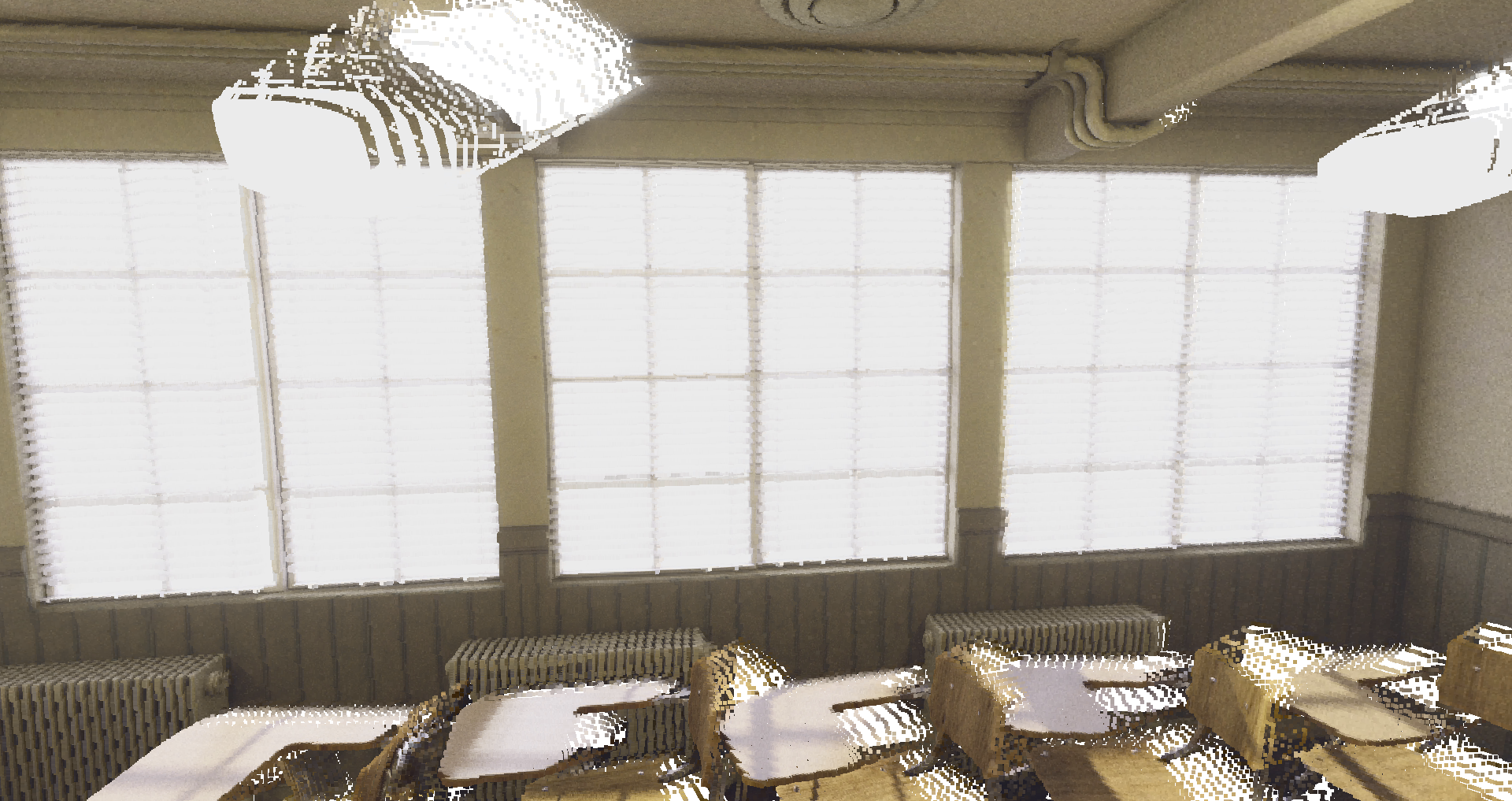} & 
    \includegraphics[width=0.235\textwidth]{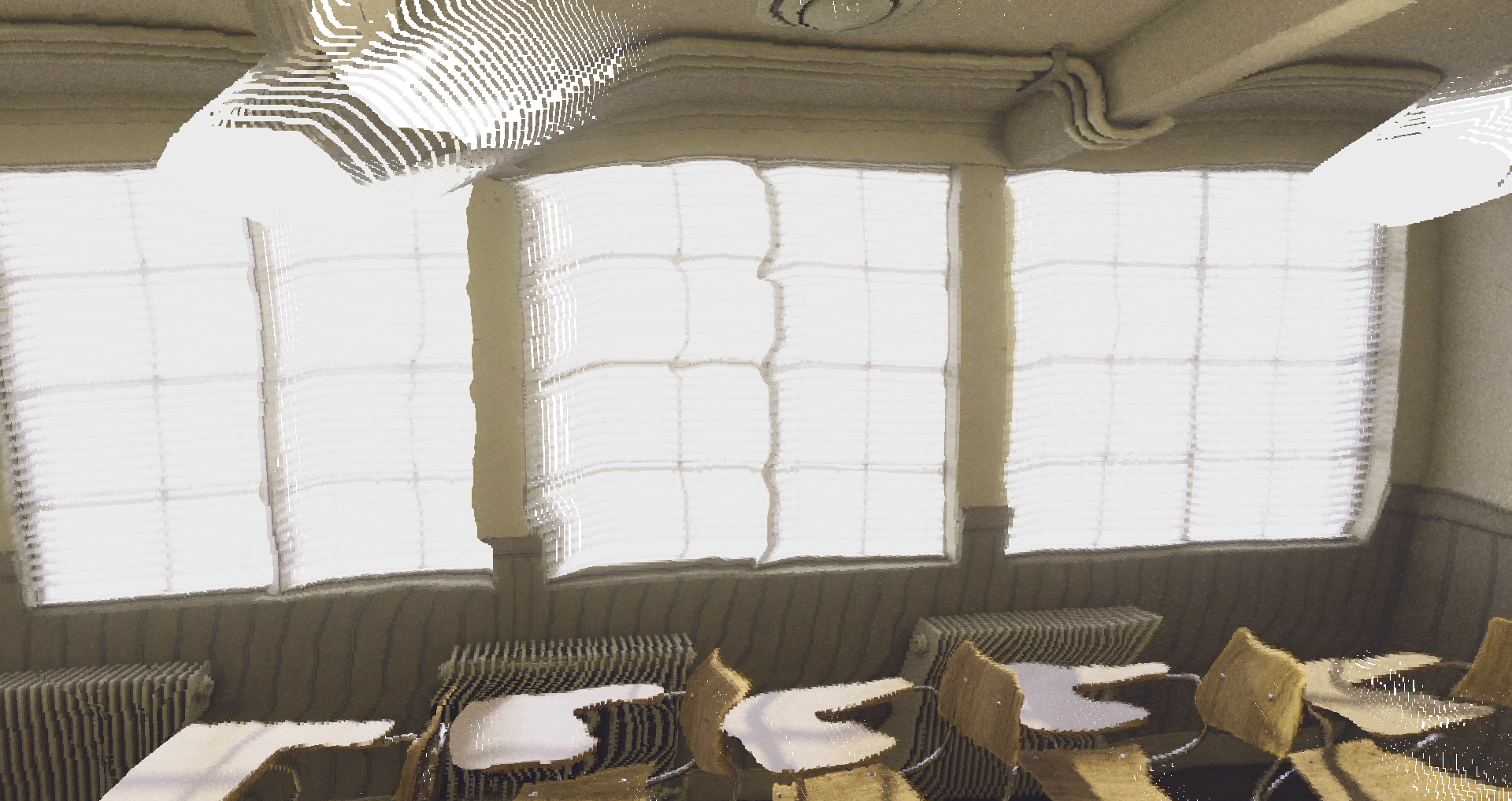} & 
    \includegraphics[width=0.235\textwidth]{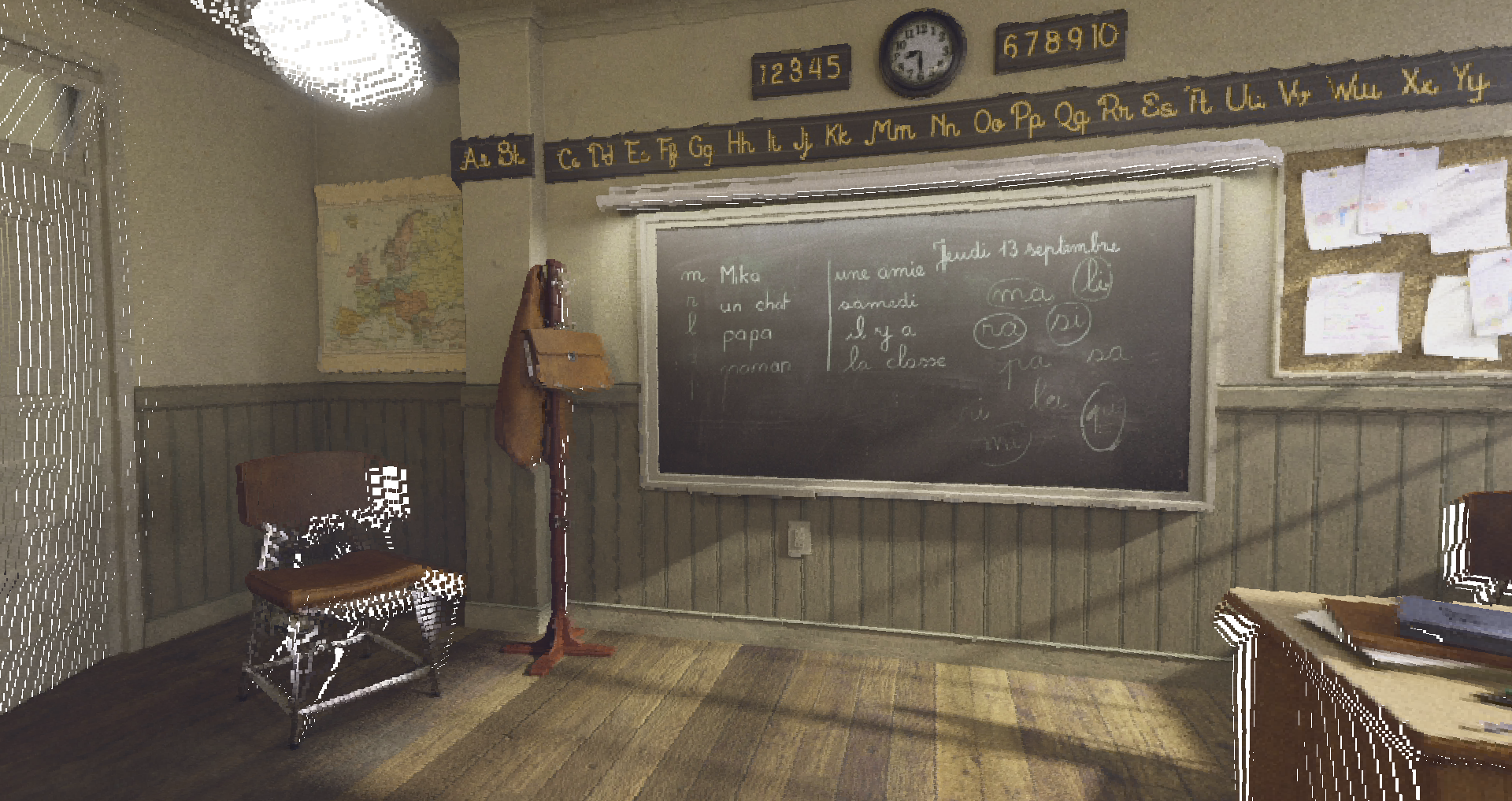} & 
    \includegraphics[width=0.235\textwidth]{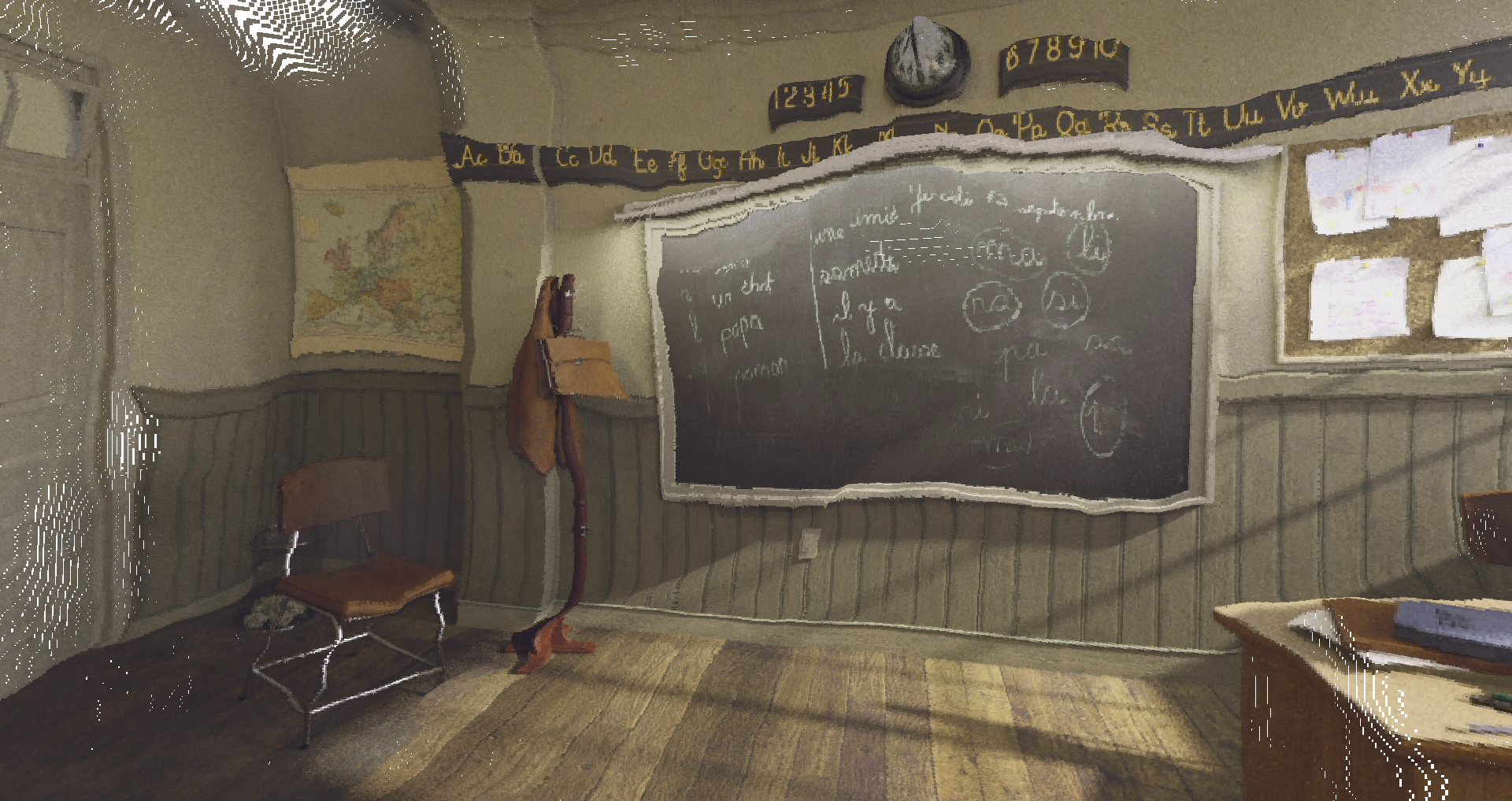}
    \\
    (a) & (b) & (c) & (d) \\
    \end{tabular}
    \caption{Example of failure cases. (a) Ground-truth depth map~(top) and 3D reconstruction~(bottom) of (b). (b) Estimated depth map and 3D reconstruction using incorrectly estimated pose. (c) Ground-truth depth map~(top) and 3D reconstruction~(bottom) of (d). (d) Estimated depth map and 3D reconstruction using the horizontally aligned cameras.} \label{fig:drawbacks}
\end{figure*}

In addition to quantitative analysis, we qualitatively validated the proposed dense depth estimation method using real-world images with a multi-camera setting. \fref{fig:result_real} shows that the estimated dense depth maps using the proposed algorithm and the 3D view of the resulting point cloud using Open3D~\cite{Zhou2018}. In the experiment, we cropped the bottom 25\% pixels of the $360$\degree\: images to remove the captured tripod. The estimated dense depth maps were smoothed using a $5 \times 5$ Gaussian filter with a mean of $0.04$ and standard deviation $1.1$ and $5 \times 5$ median filter before generating the 3D views from the point cloud~\cite{huang1979fast}.

\fref{fig:drawbacks} shows examples of some failure cases. The proposed depth estimation algorithm can possibly provide inaccurate depth in the absence of accurate pose information as shown in \fref{fig:drawbacks}(b). Although the proposed pose estimation algorithm provides a practical benefit in translation estimation, it can yield inaccurate estimates due to randomness in RANSAC~\cite{fischler1981random}. Like other conventional methods, the estimates of depth tend to be imprecise when cameras are horizontally placed as the disparity is too small, resulting in errors in depth estimation as shown in \fref{fig:drawbacks}(d). There are tiny performance differences between the choice of hyperparameters such as window size depending on the characteristics of images, as demonstrated in the ablation study of \fref{fig: window}. The optimality of hyperparameter selection in the context of depth estimation will be a subject of a future research topic. 

\section{Conclusion}
\label{sec:conclusion}
This paper developed a practical pipeline to estimate dense depth for multiview 360\degree\: images. We employed a two-view spherical camera model and extended it to multiview configurations. To improve camera pose estimation, we proposed the scaled translation scheme measuring from initial depth. In addition, we presented the dense depth estimation algorithm that minimizes photonic reprojection error using virtual depth. The experimental results verified that the proposed pipeline improves estimation accuracy compared to the state-of-art dense depth estimation methods.


%
%

\bibliographystyle{spmpsci}      
\bibliography{refs}

\end{document}